\documentclass[12pt,a4paper]{article}

\usepackage{natbib}  

\usepackage[utf8]{inputenc}
\usepackage{authblk}
\title{Concept-based Explainable Data Mining with VLM for 3D Detection}
\author[1]{Mai Tsujimoto\thanks{maitsuji@g.ecc.u-tokyo.ac.jp}}

\affil[1]{The University of Tokyo, Tokyo, Japan}

\date{}  

\usepackage{amsmath}
\usepackage{graphicx}     
\usepackage{adjustbox}    
\usepackage{booktabs}    
\usepackage{listings}     
\usepackage{color}        
\definecolor{codegreen}{rgb}{0,0.6,0}
\definecolor{codegray}{rgb}{0.5,0.5,0.5}
\definecolor{codepurple}{rgb}{0.58,0,0.82}
\definecolor{backcolour}{rgb}{0.95,0.95,0.95}

\lstdefinestyle{mystyle}{
    backgroundcolor=\color{backcolour},
    commentstyle=\color{codegreen},
    keywordstyle=\color{codepurple},
    stringstyle=\color{codegreen},
    basicstyle=\small\ttfamily,
    breakatwhitespace=false,
    breaklines=true,
    keepspaces=true,
    showspaces=false,
    showstringspaces=false,
    showtabs=false,
    tabsize=2
}

\begin{document}

\maketitle

\begin{abstract}
Rare-object detection remains a challenging task in autonomous driving systems, particularly when relying solely on point cloud data. Although Vision-Language Models (VLMs) exhibit strong capabilities in image understanding, their potential to enhance 3D object detection through intelligent data mining has not been fully explored. This paper proposes a novel cross-modal framework that leverages 2D VLMs to identify and mine rare objects from driving scenes, thereby improving 3D object detection performance. Our approach synthesizes complementary techniques such as object detection, semantic feature extraction, dimensionality reduction, and multi-faceted outlier detection into a cohesive, explainable pipeline that systematically identifies rare but critical objects in driving scenes. By combining Isolation Forest and t-SNE-based outlier detection methods with concept-based filtering, the framework effectively identifies semantically meaningful rare objects. A key strength of this approach lies in its ability to extract and annotate targeted rare object concepts such as construction vehicles, motorcycles, and barriers. This substantially reduces the annotation burden and focuses only on the most valuable training samples. Experiments on the nuScenes dataset demonstrate that this concept-guided data mining strategy enhances the performance of 3D object detection models while utilizing only a fraction of the training data, with particularly notable improvements for challenging object categories such as trailers and bicycles compared with the same amount of random data. This finding has substantial implications for the efficient curation of datasets in safety-critical autonomous systems.
\end{abstract}

\section{Introduction}
Autonomous driving systems rely heavily on an accurate perception of the surrounding environment, and 3D object detection is a critical component of this perception stack \cite{liu2023bevfusion,yin2021center,lang2019pointpillars,bai2022transfusion,mao20233d,guo2020deep,QIAN2022108524}. Although common objects, such as cars and pedestrians, are well-represented in training datasets, rare objects or unusual scenarios pose significant challenges owing to their limited occurrence in the collected data.

Traditional approaches to improve detection performance often rely on collecting and annotating additional data, which is time-consuming and expensive, particularly for rare events. Moreover, simply increasing the dataset size without considering the diversity of scenarios may not effectively address the long-tail distribution problem inherent in real-world driving scenes. This is evident in widely used autonomous driving datasets such as nuScenes \cite{caesar2020nuscenes}, where the class distribution is highly imbalanced, with cars (493,322 instances) and pedestrians (208,240 instances) far outnumbering construction vehicles (14,671 instances) and bicycles (11,859 instances). This imbalance creates significant challenges for developing robust detection systems.

Vision-language models (VLMs) have demonstrated remarkable capabilities in understanding and describing visual content. These models enable applications such as image captioning, visual question answering, and zero-shot classification~\cite{radford2021learning,liu2024grounding,li2022blip,liu2023visual,bommasani2021opportunities,kojima2022large}. VLMs offer a promising approach for analyzing and categorizing driving scenes based on their semantic content.

This paper introduces a \textbf{concept-based explainable data mining framework} (Fig~\ref{fig:concept_mining_flow}) that leverages Vision-Language Models (VLMs) to identify and select valuable training samples for 3D object detection. VLMs uniquely connect visual features with linguistic concepts and enable semantic understanding of complex objects beyond what simple shape features can capture. This is particularly effective for rare object detection, which is challenging when using point cloud data alone. Our framework comprises three main components:
\begin{itemize}
    \item \textbf{Object Concept Embedding} system detects objects in 2D images, generates captions using VLMs, and extracts feature embeddings to create semantic representations of each object.
    \item \textbf{Rare Object Mining} method employs dimensionality reduction through t-SNE \cite{maaten2008visualizing} and outlier detection algorithms \cite{liu2008isolation} to identify unusual objects or scenarios that deviate from common distributions.
    \item \textbf{Targeted Data Mining} framework selects scenes containing specific rare object concepts for annotation and inclusion in the training dataset, thereby improving the detection performance for under-represented classes.
\end{itemize}

This work contributes a novel framework for \textbf{explainable data mining} that leverages Vision-Language Models to identify and select valuable training samples for 3D object detection. This transparency enhances the reliability of decision criteria in safety-critical autonomous driving systems. Our concept-based approach enables \textbf{targeted mining} for specific object categories and focuses improvement efforts on challenging classes. Experiments demonstrate substantial performance enhancement for rare object categories while using only \textbf{20\%} of the training data compared with random sampling. We provide both qualitative and quantitative analyses of the detected rare objects and offer insights into the types of scenarios that contribute most to performance improvement. The approach achieves a substantial reduction in annotation costs and intelligently extracts only scenes with targeted safety-critical concepts such as traffic cones, bicycles in unusual positions, and temporary road infrastructure, which leads to more efficient development of robust autonomous driving systems.

\textbf{Terminology and scope.} In this paper, we use the term \emph{rare objects} to refer to safety-relevant object classes or scene configurations that appear infrequently in the training distribution. Unless otherwise stated, we consistently use \emph{rare objects}. When we deliberately focus on a small subset of classes (e.g., bicycles and motorcycles), we call them \emph{target classes}. We use \emph{uncommon concepts} to describe VLM-extracted concepts that are outside the common class list. In our pipeline, these are treated as rare objects for data selection.

For clarity, 2D concept mining is used only for scene selection. The 3D detector is solely trained and evaluated on 3D annotations from the selected scenes, which creates a concise 2D \(\rightarrow\) data selection \(\rightarrow\) 3D training loop.

\section{Related Work}
\textbf{3D object detection.}
In recent years, substantial advancements have been made in 3D object detection, with various approaches tailored to process point cloud data.
PointPillars~\cite{lang2019pointpillars} introduced an efficient encoder for converting point clouds into a pseudo-image format that can be processed using standard 2D convolutional networks. CenterPoint~\cite{yin2021center} proposed a center-based representation for 3D objects that simplifies the detection pipeline and improves performance. For multimodal fusion, TransFusion~\cite{bai2022transfusion} utilized transformer decoders to perform robust fusion of LiDAR and camera data through soft association, whereas BEVFusion~\cite{liu2023bevfusion} presented a unified bird's-eye view representation for multitask multisensor fusion that achieves state-of-the-art performance.

\textbf{Vision-Language Models.}
Vision-language models (VLMs) are an effective means of connecting visual and textual information. CLIP~\cite{radford2021learning} demonstrated impressive zero-shot capabilities by training on a large dataset of image-text pairs, establishing a robust foundation for image-text understanding. BLIP~\cite{li2022blip} enhances image-text pre-training with bootstrapped language-image pre-training, improving performance on various vision-language tasks. More recent models have extended these capabilities, with GroundDINO~\cite{liu2024grounding} enabling grounded detection tasks and LLaVA~\cite{liu2023visual} incorporating large language models to enhance visual reasoning capabilities. This work leverages these advances in VLMs, particularly CLIP for feature extraction and Qwen2-VL~\cite{wang2024qwen2} for image captioning, to create semantic representations of objects for data mining.

\textbf{Data Mining for 3D Object Detection.}
Several approaches have been proposed for data mining in the context of object detection. 
For 2D detection, curriculum learning~\cite{bengio2009curriculum} was originally introduced in classification tasks to improve convergence by gradually increasing the sample difficulty and has since been extended to detection scenarios. 
In contrast, hard example mining strategies specifically designed for detection, such as Online Hard Example Mining (OHEM)~\cite{shrivastava2016training} for two-stage detectors and Focal Loss~\cite{lin2018focal} for one-stage detectors, focus on prioritizing challenging samples during training to improve robustness. 
In the context of 3D detection, Jiang et al.~\cite{jiang2022improving} addressed the intra-class long-tail problem by mining rare examples through density estimation in the feature space using a normalizing flow model. Their approach selects and augments underrepresented instances in point cloud data, leading to significant improvements in the detection performance for rare objects. 
Peri et al.~\cite{peri2023towards} developed methods that emphasize rare examples to improve the accuracy of long-tailed 3D detection. 
Lu et al.~\cite{lu2023open} enhanced detection capabilities without relying on 3D annotations through open-vocabulary learning. 
VLMine~\cite{ye2025vlmine} explored the use of vision-language models (VLMs) for long-tail data mining in 3D object detection.

\textbf{Concept Bottleneck Models and Label-Free Interpretability.}
Concept Bottleneck Models (CBMs)~\cite{koh2020concept} have emerged as a powerful approach for enhancing the interpretability of deep learning models. CBMs first predict human-interpretable concepts and then use these concepts to predict the final output, enabling the direct manipulation of concepts for interventions and explanations. Building on this foundation, the Label-Free Concept Bottleneck Model (LF-CBM)~\cite{oikarinen2023label} transforms neural networks into interpretable CBMs without requiring labeled concept data, leveraging large language models for concept generation and CLIP for concept extraction. Further advancing this field, CEIR~\cite{cui2023ceir} integrates CBMs into unsupervised representation learning with GPT-4 and Variational Autoencoders to enable human-centric explanations without labeled data.

\textbf{Foundation Models and Object Detection.}
Foundation models~\cite{bommasani2021opportunities}, pretrained on vast amounts of data, have revolutionized various computer vision tasks, including object detection, by providing powerful transferable representations. Large language models like GPT-4~\cite{achiam2023gpt} have demonstrated remarkable capabilities in understanding and generating text, and are increasingly being integrated with vision components to create powerful multimodal systems. In the context of object detection, efficient object detectors such as YOLOX~\cite{ge2021yolox} and YOLOv8~\cite{reis2023realtime} offer strong performance with real-time capabilities and can serve as base models for concept-enhanced frameworks.

\textbf{The proposed approach.}
We introduce a concept-based approach that integrates VLMs with a novel data mining framework that identifies rare objects through semantic content rather than just geometric features. The proposed method extends concept-based visualization and interpretability techniques to 3D object detection and enhances explainability in data mining while improving detection of semantically interesting but rare objects with limited training data. The proposed approach leverages foundation models for concept extraction and combines them with efficient detection architectures. This enables a more nuanced understanding of object relationships in complex scenes and creates an explainable framework that improves rare object detection and reduces annotation costs.

\section{Method}
The proposed concept-based explainable data mining framework consists of three main components. We first extract object concept embeddings, then perform rare object mining, and finally apply targeted data mining for 3D detection. 

\begin{figure}[t]
\centering
\includegraphics[width=0.77\linewidth]{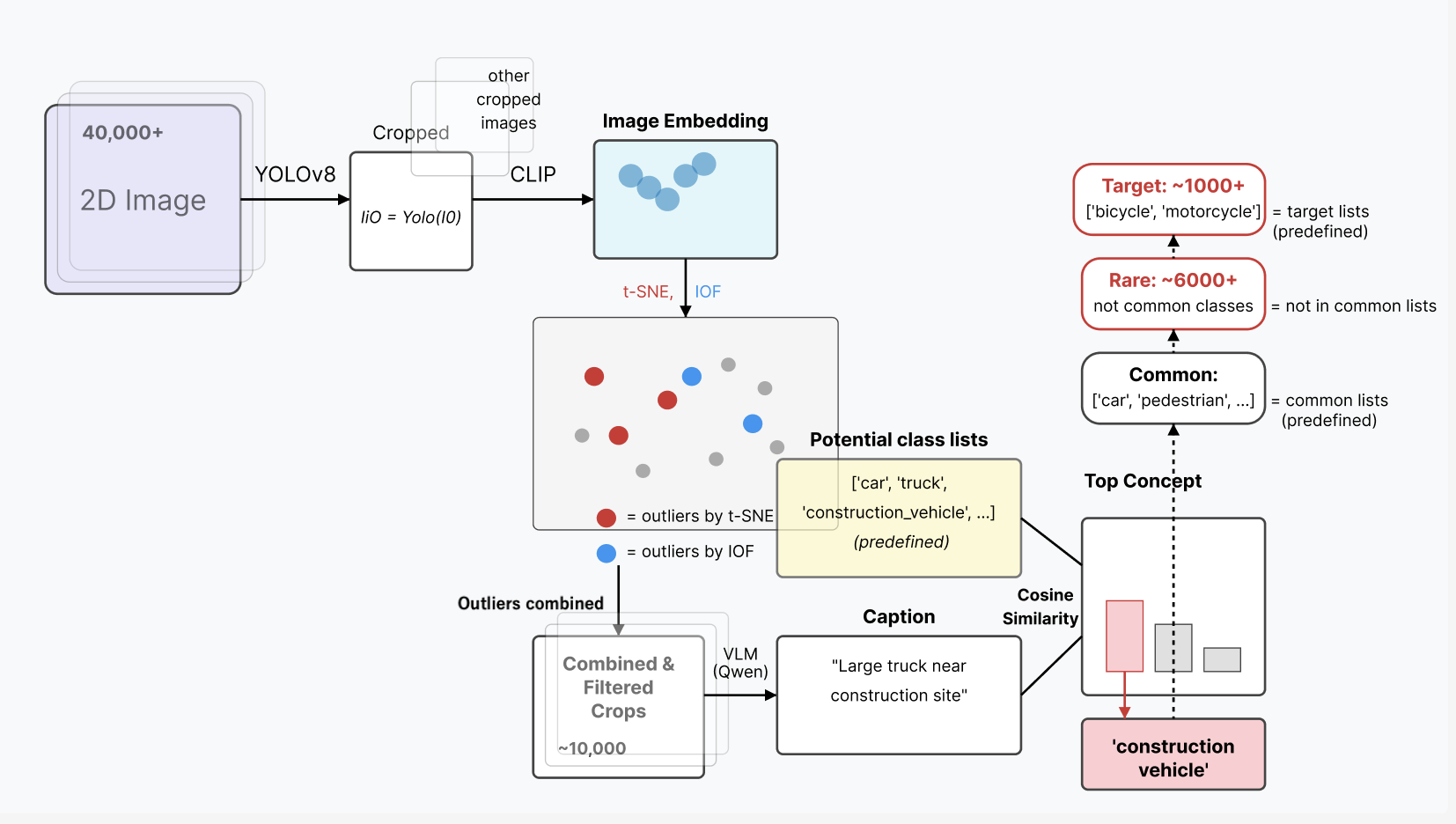}
\caption{\small 
Overview of the refined targeted concept mining framework used for selecting rare and safety-critical objects from large-scale 2D image datasets. YOLOv8 was first used to detect and crop objects from over 40,000 images. CLIP is then applied to extract image embeddings, which are analyzed using both t-SNE and Isolation Forest to identify outliers. Approximately 10,000 detected outlier crops were passed to a vision-language model (Qwen2-VL) to generate captions. Each caption is matched to predefined class concepts using cosine similarity in the CLIP embedding space. Detected objects are categorized into three groups based on the top-matching concept: Target (~1,000; e.g., 'bicycle', 'motorcycle'), Rare (~6,000; classes not in the common list; e.g., not in ['car', 'pedestrian']), and Common (e.g., 'car', 'pedestrian'). The final training dataset was constructed by including all images with target objects and random sampling from the rare set, forming a "Random 10\% + Target 10\%" strategy that enhanced rare class representation while maintaining dataset balance.
} 
\label{fig:concept_mining_flow}
\end{figure}

\subsection{Object Concept Embedding}

The object concept embedding pipeline (Fig~\ref{fig:concept_mining_flow}) transforms 2D images into a semantic space in which objects can be compared based on their visual and conceptual properties.

\textbf{Object Detection and Cropping.} YOLOv8 was applied to detect and crop objects from the input images, as shown on the left side of Fig.~\ref{fig:concept_mining_flow}. This process extracts object images from over 40,000 2D images through $I^O_i = \text{YOLOv8}(I_i)$, where $I_i$ is the input image and $I^O_i$ is the set of cropped object images.

\textbf{Feature Extraction.} For each cropped object, CLIP~\cite{radford2021learning} extracts feature embeddings $F_i = \text{CLIP}(I^O_i)$. These features correspond to the "Image Embedding" component in the figure and are used for the subsequent analysis. We apply two outlier detection methods to these embeddings. Isolation Forest~\cite{liu2008isolation} identifies objects that deviate from the common distribution through $O^{IF}_i = \text{IsolationForest}(F_i) \in \{1, 0\}$, where $1$ indicates an outlier and $0$ indicates an inlier. Meanwhile, t-SNE~\cite{maaten2008visualizing} projects the embeddings to 2D space as $L_i = \text{t-SNE}(F_i)$, followed by proximity-based outlier detection computed as $D_i = \frac{1}{k}\sum_{j \in \text{kNN}(L_i, k)} \|L_i - L_j\|_2$ and $O^{tsne}_i = \mathbf{1}(D_i > \tau_{tsne})$, where $\mathbf{1}(\cdot)$ is the indicator function that returns 1 if the distance exceeds the threshold $\tau_{tsne}$ and 0 otherwise.

\textbf{Outlier Combination.} The outliers from both methods are combined as shown in the "Outliers combined" section of the figure using $O_i = 2 \cdot O^{tsne}_i + O^{IF}_i$. This combined approach detects diverse anomaly patterns that cannot be fully captured by a single outlier detection method. Specifically, t-SNE excels at clustering visually similar elements and identifies outliers in local neighborhoods, while Isolation Forest demonstrates strength in detecting global nonlinear anomaly patterns through recursive partitioning. Compared to alternative combinations such as LOF+IOF, this approach contributes more effectively to visualization and explainability while empirically producing superior results, as detailed in our experimental trials (see Appendix). 

However, this approach has inherent limitations. T-SNE is non-parametric and sensitive to variations in hyperparameters and sampling density. Therefore, we use the combined score to propose candidates for concept filtering, rather than for making final data selection decisions. Although alternative outlier detectors could be employed (e.g., Local Outlier Factor, one-class SVM, and isolation-based variants), we selected the t-SNE and Isolation Forest combination because of their complementary detection behaviors and enhanced visual interpretability in the embedding space.

\textbf{Concept-based Filtering.} The detected outliers were filtered based on their concepts to focus on meaningful rare objects.

\begin{equation}
R_i = 
\begin{cases}
1, & \text{if } O_i > 0 \text{ and } C_i \cap \text{CommonClasses} = \emptyset \\
0, & \text{otherwise}
\end{cases}
\end{equation}

where $\text{CommonClasses}$ is a set of common object classes that are well represented in the dataset.

\textbf{Concept Extraction.} For the identified outlier samples, the Qwen2-VL~\cite{wang2024qwen2} vision-language model generates detailed captions as $W_i = \text{Qwen2-VL}(I^O_i)$. The generated captions are compared with a predefined concept list to extract relevant object concepts through \\$C_i = \text{ParseConcepts}(W_i, \text{ConceptList})$.
As shown on the right side of the figure, this categorizes objects into three groups. The `Target' group includes approximately 1,000 objects that match predefined target classes, such as `bicycle' and `motorcycle.' The `Rare' group includes approximately 6,000 objects whose concepts are not in the common class list. The `Common' group includes objects that match common classes.

We optimized the concept embedding approach by making several enhancements. The captioning process incorporates class information detected by YOLOv8, which provides the VLM with prior knowledge of the detected object. Specialized prompting strategies direct the Qwen2-VL-2B-Instruct model to concentrate on features relevant to driving. We developed a hierarchical concept taxonomy that extends beyond the standard object classes in the nuScenes dataset. CLIP's semantic embedding space computes the similarities between the generated captions and the predefined concept vocabulary.

\subsection{Data Mining for 3D Detection}

\textbf{Data Mining for Target Label Concept.}
This approach focuses on specific object categories of interest such as motorcycle or bicycle. Scenes that contain objects whose top concept matches the target concept are selected through \\$S_{target} = \{s_i | \exists o \in s_i \text{ such that } \text{TopConcept}(o) = \text{TargetConcept}\}$. These scenes were then annotated for 3D detection and added to the training dataset.

\textbf{Data Mining for Rare Objects.}
This approach improves the overall robustness to unusual scenarios by selecting scenes that contain objects identified as rare through \\$S_{rare} = \{s_i | \exists o \in s_i \text{ such that } R_o = 1\}$. Scenes are filtered based on the top concept of the detected objects and prioritize those with uncommon concepts. The selected scenes were annotated for 3D detection and added to the training dataset.

We further explored two variant approaches. \textbf{Random-Target+} incorporates potential target classes into the pipeline regardless of their outlier status, while \textbf{Random-Target} employs a more selective filtering strategy of Fig.~\ref{fig:concept_mining_flow}. These approaches are described in detail in Appendix A.1.

\section{Experiments}

\subsection{Experimental Setup}

We used YOLOv8-L as a 2D object detection backbone. For semantic representation, we extracted CLIP (ViT-B/32) embeddings, which were subsequently processed by an Isolation Forest with a contamination parameter of 0.2 for outlier filtering and t-SNE for dimensionality reduction. Qwen2-VL-2B-Instruct then utilized the resulting embeddings to generate descriptive captions.
To assess the 3D detection performance, we evaluated the nuScenes test set \cite{caesar2020nuscenes} using CenterPoint \cite{yin2021center} as the baseline detector.
For the final dataset construction, we adopted a "Random 10\% + Target 10\%" sampling strategy. All images that contained predefined target objects were included, and an additional 10\% subset was randomly sampled from the rare set, along with a separate non-overlapping random 10\% from the entire dataset. This strategy ensures sufficient coverage of safety-critical rare classes and preservation of the overall distributional balance.

\textbf{Evaluation protocol and baselines.}
All reported improvements were measured against random sampling baselines with matched data volumes. The models were trained using the same 3D detector and recipe under identical hyperparameters, and the targeted and random subsets were constructed to be non-overlapping. The metrics were reported on the nuScenes test set. A head-to-head comparison with hard example mining, active learning, and other data selection strategies is left for future work.

\subsection{Results}

\textbf{Data Mining Results.}

To evaluate the effectiveness of the data mining approach, we compared the following training strategies:
\begin{itemize}
    \item Random 10\%: Randomly selects 10\% of the training data
    \item Random 20\%: Randomly selects 20\% of the training data
    \item Full 100\%: Uses the entire training dataset
    \item Random-Rare: Combines 10\% random data with 10\% data mined with the rare object strategy
    \item Random-Target: Combines 10\% random data with 10\% data mined with the target concept strategy and focuses on bicycles and motorcycles
\end{itemize}

Table \ref{tab:results} presents the performance results. Our approach yielded substantial improvements, particularly in rare and challenging categories. The Random-Target strategy improved mAP from 43.3\% to 44.7\%, a gain of 1.4 points over Random 20\%. Notable improvements appeared in targeted classes, with motorcycle detection improving from 31.1\% to 34.6\% and bicycle detection advancing from 7.5\% to 11.2\%. These results validate our approach of leveraging VLM understanding for targeted data mining.

\begin{table*}[t]
\small 
\centering
\caption{\small
Comparison of detection performance on the nuScenes test set measured by Average Precision (AP). Abbreviations denote Construction Vehicle (CV), Trailer (Tra), Motorcycle (Mot), Pedestrian (Ped), Bicycle (Bic), and Traffic Cone (Traf). Random-Rare combines 10\% random data with 10\% rare-class data, while Random-Target combines 10\% random data with 10\% data focused on target classes (Motorcycle and Bicycle). Asterisks (*) indicate improvements that exceed 0.5 AP compared to Random 20\%. Random-Target shows clear improvements in target classes, with AP gains of +3.5 for motorcycles and +3.7 for bicycles over the Random 20\% baseline. Random-Rare yields meaningful gains in rare classes where APs in Random 10\% and Random 20\% are low, despite using the same total data volume as Random 20\%. Random-Target achieves a 1.4 point mAP increase over Random 20\%, which underscores the benefit of targeted data sampling to enhance detection performance, particularly for underrepresented rare classes.}
\label{tab:results}
\begin{adjustbox}{width=\textwidth}
\begin{tabular}{lcccccccccccc}
\toprule
\textbf{Data} & \textbf{mAP} & \textbf{Car} & \textbf{Truck} & \textbf{Bus} & \textbf{Tra} & \textbf{CV} & \textbf{Ped} & \textbf{Mot} & \textbf{Bic} & \textbf{Traf} & \textbf{Barrier} \\
\midrule
Random 10\% & 39.6 & 79.3 & 43.0 & 54.7 & 26.1 & 5.9 & 68.9 & 24.8 & 4.3 & 41.5 & 47.7 \\
Random 20\% & 43.3 & 81.1 & 45.2 & 59.7 & 28.2 & 8.8 & 73.2 & 31.1 & 7.5 & 46.4 & 52.4 \\
Full 100\% (reference) & 56.3 & 84.9 & 53.8 & 66.2 & 31.8 & 15.7 & 83.6 & 55.3 & 37.4 & 66.6 & 67.5 \\
Random-Rare (ours) & 44.0* & 81.4 & 46.1* & 59.9 & 30.5* & 9.4* & 73.0 & 31.2* & 7.8 & 48.5* & 52.4 \\
Random-Target (ours) & 44.7* & 81.2 & 45.5* & 58.1 & 31.2* & 9.3* & 72.9 & \textbf{34.6*} & \textbf{11.2*} & 48.7* & 54.1* \\
\bottomrule
\end{tabular}
\end{adjustbox}
\end{table*}

Our framework effectively identifies and extracts rare objects. The success in both motorcycle detection at 34.6\% AP and bicycle detection at 11.2\% AP demonstrates how the VLM's accurate understanding of these categories enables precise data selection. This targeted approach proves more effective than general rare object mining, as the Random-Target strategy outperforms both Random 20\% and Random-Rare approaches to enhance specific objects in 3D object detection.

Performance analysis of other object categories, such as construction vehicles, provides additional insights into the behavior of our framework. A detailed analysis of construction vehicle detection challenges and opportunities is provided in Appendix (Section A.2). 

The framework also achieves strong efficiency. By reaching 80\% of the full dataset performance while using only 20\% of the data, it offers practical value for real-world applications where annotation resources are limited. This efficiency proves particularly valuable for autonomous driving applications, where detecting certain vulnerable road users is more critical.

\textbf{Analysis.}

\begin{figure}[t]
\centering
\includegraphics[width=0.4\linewidth]{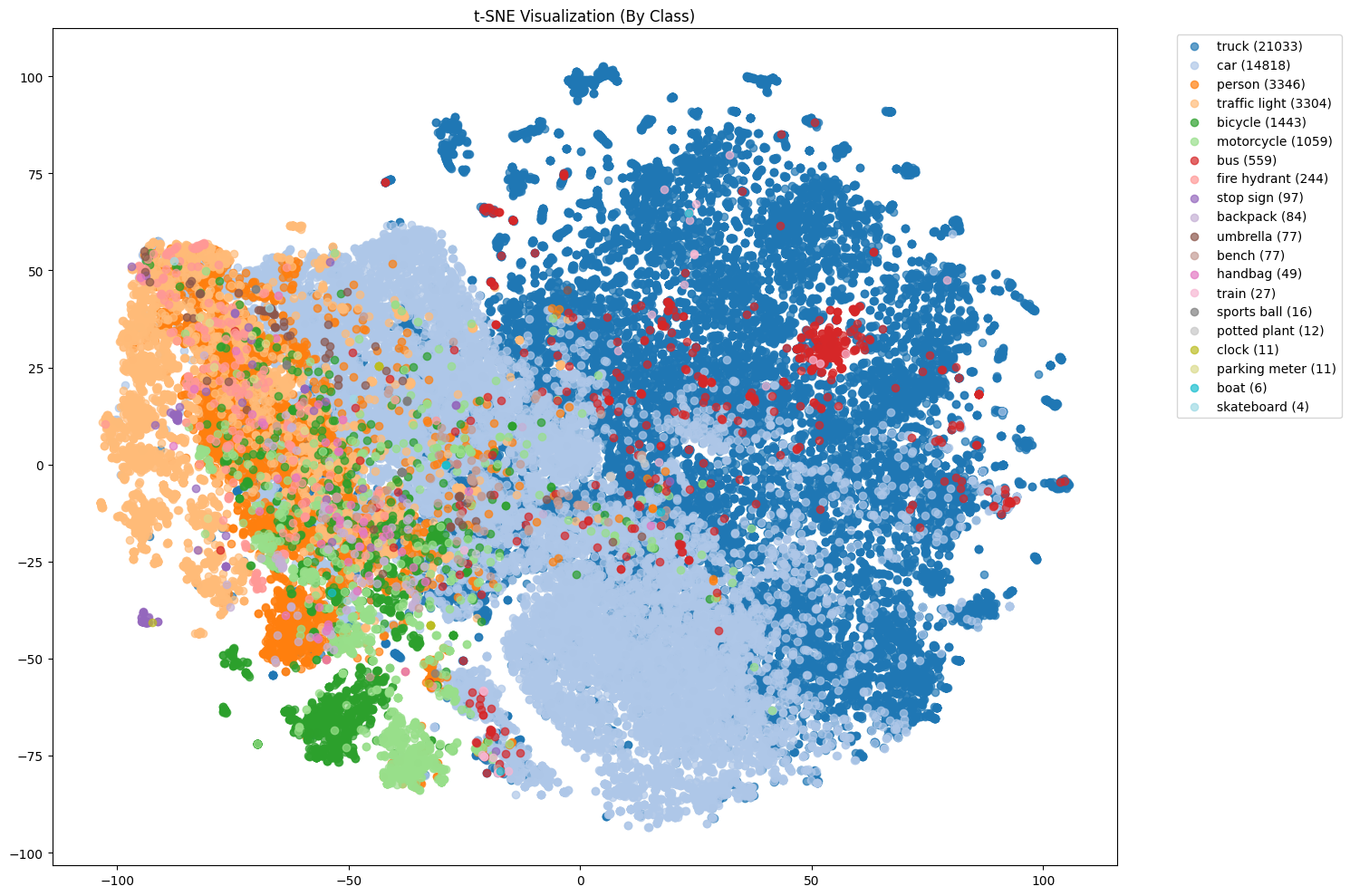}
\includegraphics[width=0.4\linewidth]{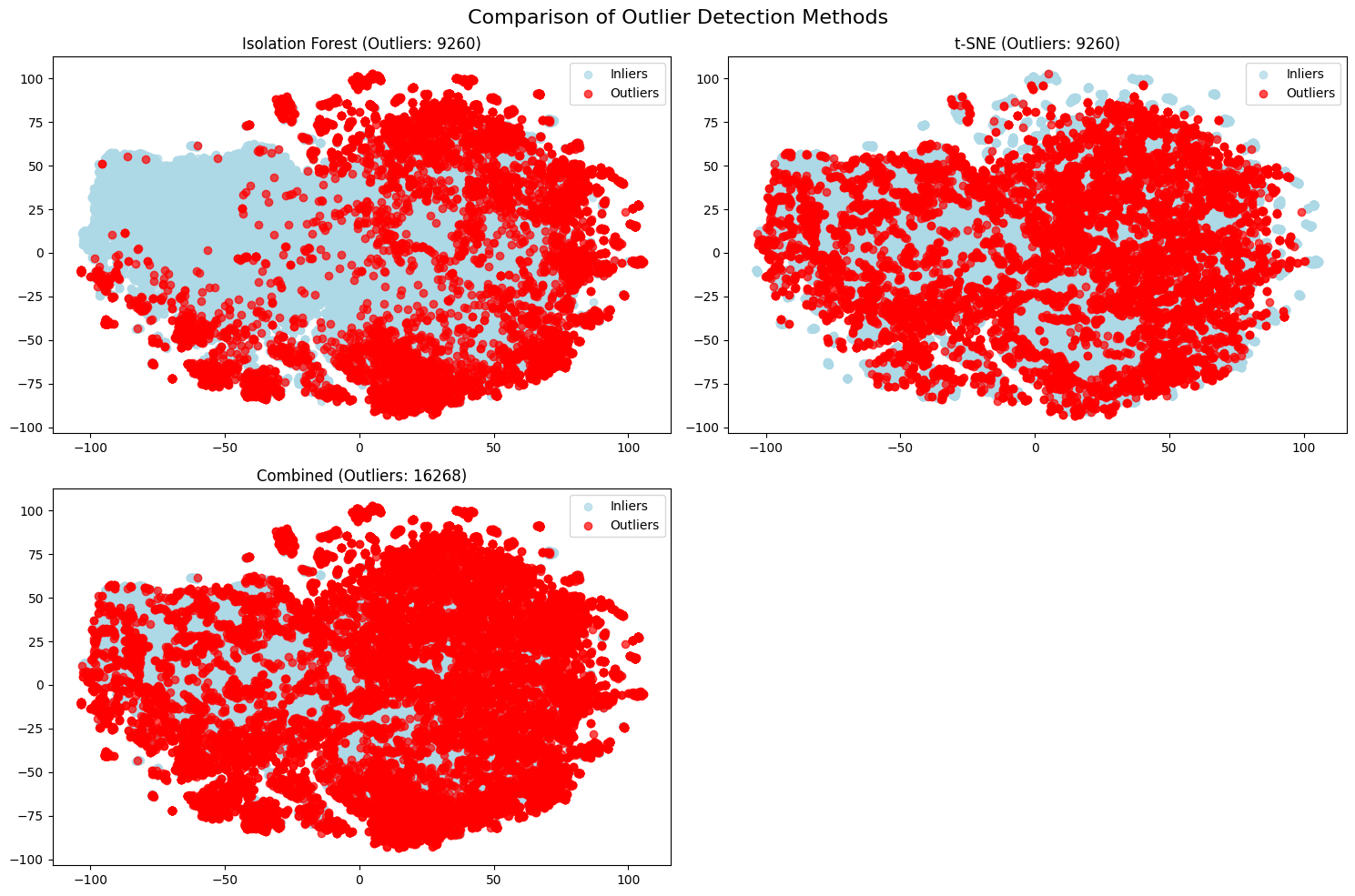}
\caption{\small t-SNE visualization of object embeddings and outlier detection results. 
(\textit{Left}) Embeddings colored according to object category. 
(\textit{Top-left}) Outliers detected using the Isolation Forest. 
(\textit{Top-right}) Outliers based on t-SNE anomaly regions. 
(\textit{Bottom-left}) Combined outliers detected by both methods. 
The red points indicate anomalous samples, and the blue points indicate inliers. (The red points appear to dominate the plots due to overplotting, but in fact only about 20-30\% of the samples were detected as outliers.)}
\label{fig:tsne}
\end{figure}

To better understand the distribution of object features and evaluate our outlier detection strategy, we visualized the learned embeddings using t-SNE.
Figure \ref{fig:tsne} presents visualizations of the nuScenes dataset.
The left image shows embeddings colored by object category and demonstrates how different object classes form distinct clusters in embedding space.
The three images on the right illustrate the outlier detection results.
The top-left shows outliers detected solely by Isolation Forest, the top-right shows outliers based on t-SNE anomaly regions, and the bottom-left shows the combined anomaly regions from both methods.
In all three cases, the red points represent the identified outliers.
This visualization shows that our approach identifies anomalies across the feature space, particularly in regions corresponding to rare classes, thereby validating our anomaly detection strategy.

\begin{figure}[t]
\centering
\begin{minipage}[t]{0.3\linewidth}
\centering
\includegraphics[width=0.4\linewidth]{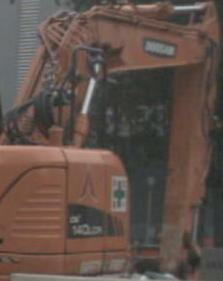}
\includegraphics[width=0.8\linewidth]{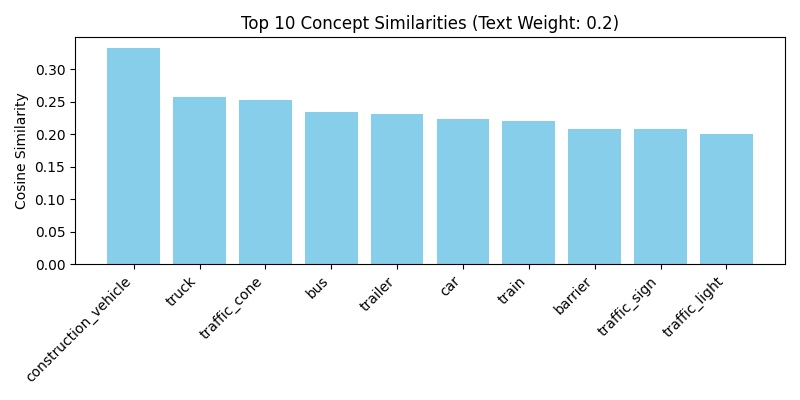}
\caption{\small An example of a detected construction vehicle, correctly identified with high concept similarity to "construction\_vehicle".}
\label{fig:construction}
\end{minipage}%
\hspace{0.01\linewidth}
\begin{minipage}[t]{0.3\linewidth}
\centering
\includegraphics[width=0.3\linewidth]{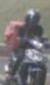}
\includegraphics[width=0.8\linewidth]{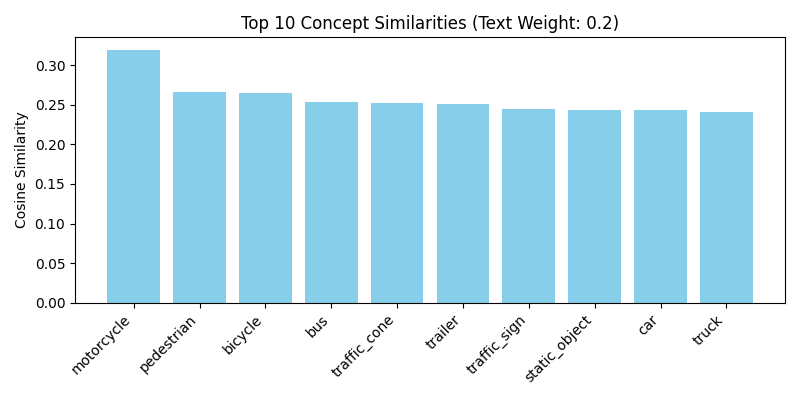}
\caption{\small Example of a motorcycle detected by the system, showing a strong association with "motorcycle" in the concept analysis.}
\label{fig:motorcycle}
\end{minipage}%
\hspace{0.01\linewidth}
\begin{minipage}[t]{0.3\linewidth}
\centering
\includegraphics[width=0.45\linewidth]{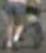}
\includegraphics[width=0.8\linewidth]{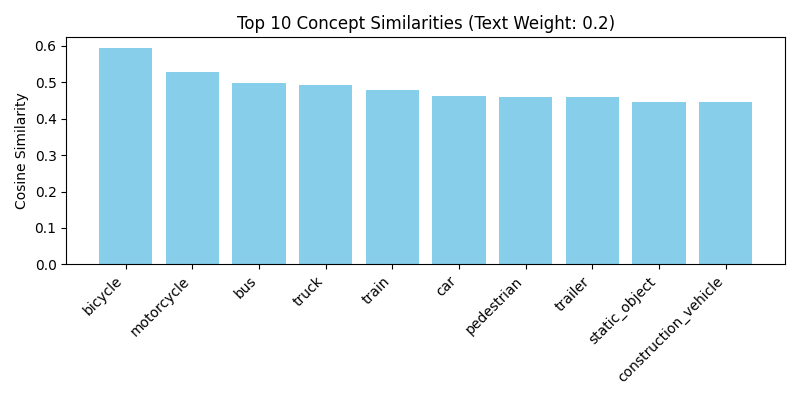}
\caption{\small Example of a bicycle detected by the system, showing strong association with "bicycle" in the concept analysis.}
\label{fig:bicycle}
\end{minipage}
\end{figure}

Our data mining framework effectively identifies various rare and challenging object categories, as illustrated in Figures \ref{fig:construction}, \ref{fig:motorcycle}, and \ref{fig:bicycle}. Figure \ref{fig:construction} shows an example of a construction vehicle detected by our system, with the bar graph visualization showing high concept similarities to "construction\_vehicle."
Similarly, Figure \ref{fig:motorcycle} demonstrates motorcycle detection with a strong association with the "motorcycle" concept, and Figure \ref{fig:bicycle} shows a bicycle with a clear concept association.
These visualizations demonstrate the success of our object detection and provide explainability through concept similarity graphs.
This explainability proves crucial when we target specific classes and allows users to understand why certain objects are selected for a training dataset.

\begin{figure}[t]
\centering
\includegraphics[width=0.48\linewidth]{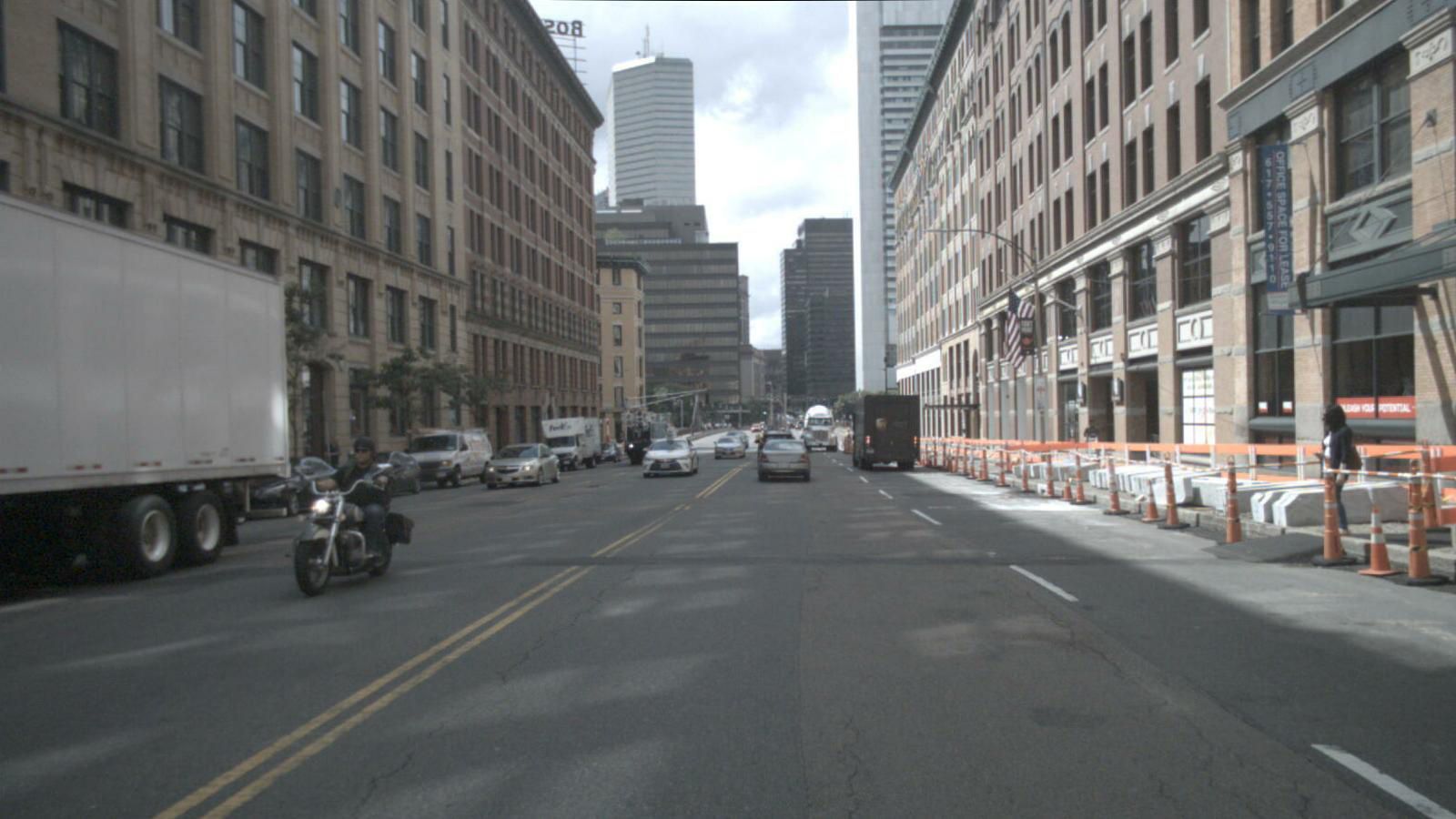}
\includegraphics[width=0.48\linewidth]{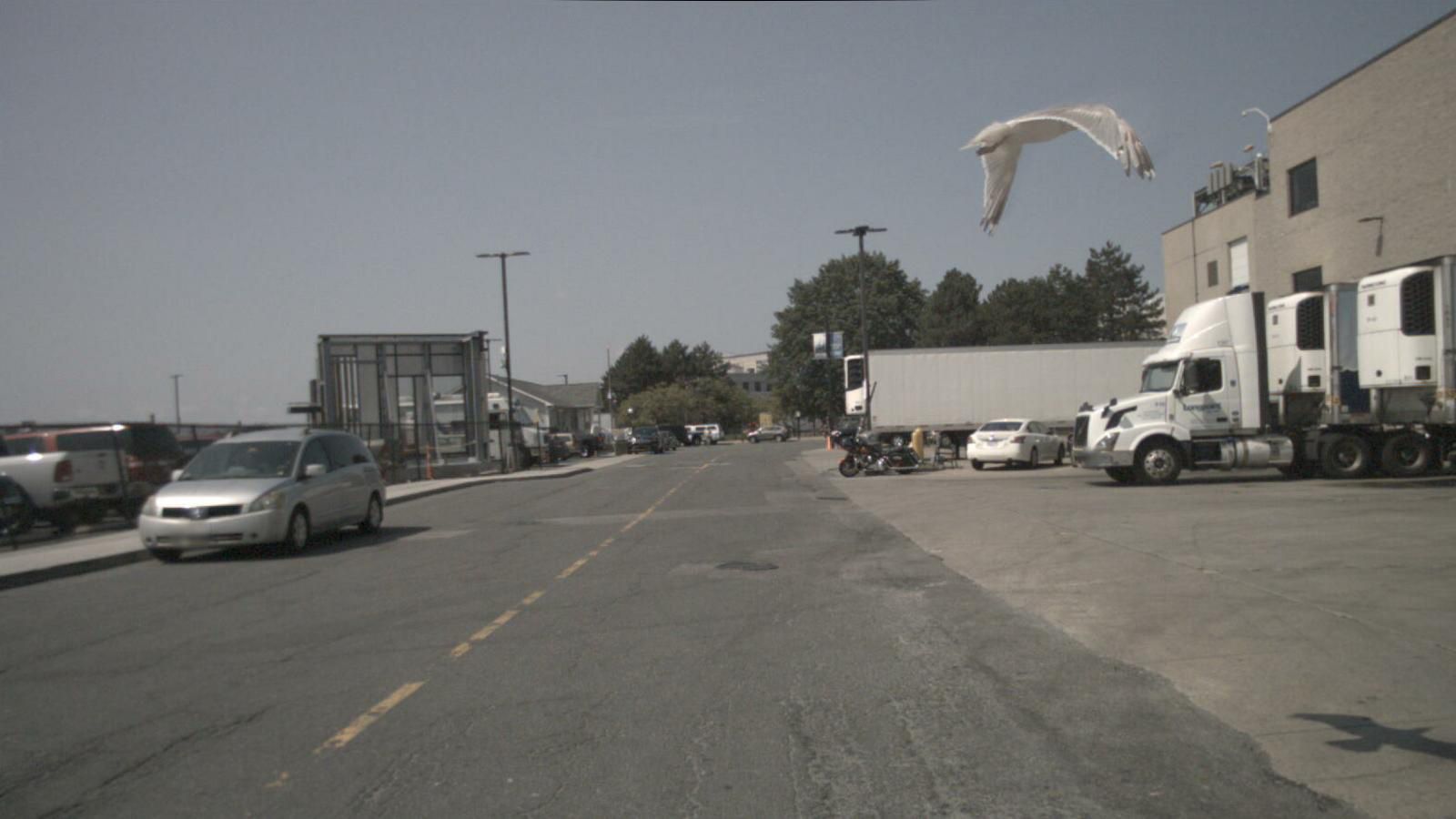}
\includegraphics[width=0.48\linewidth]{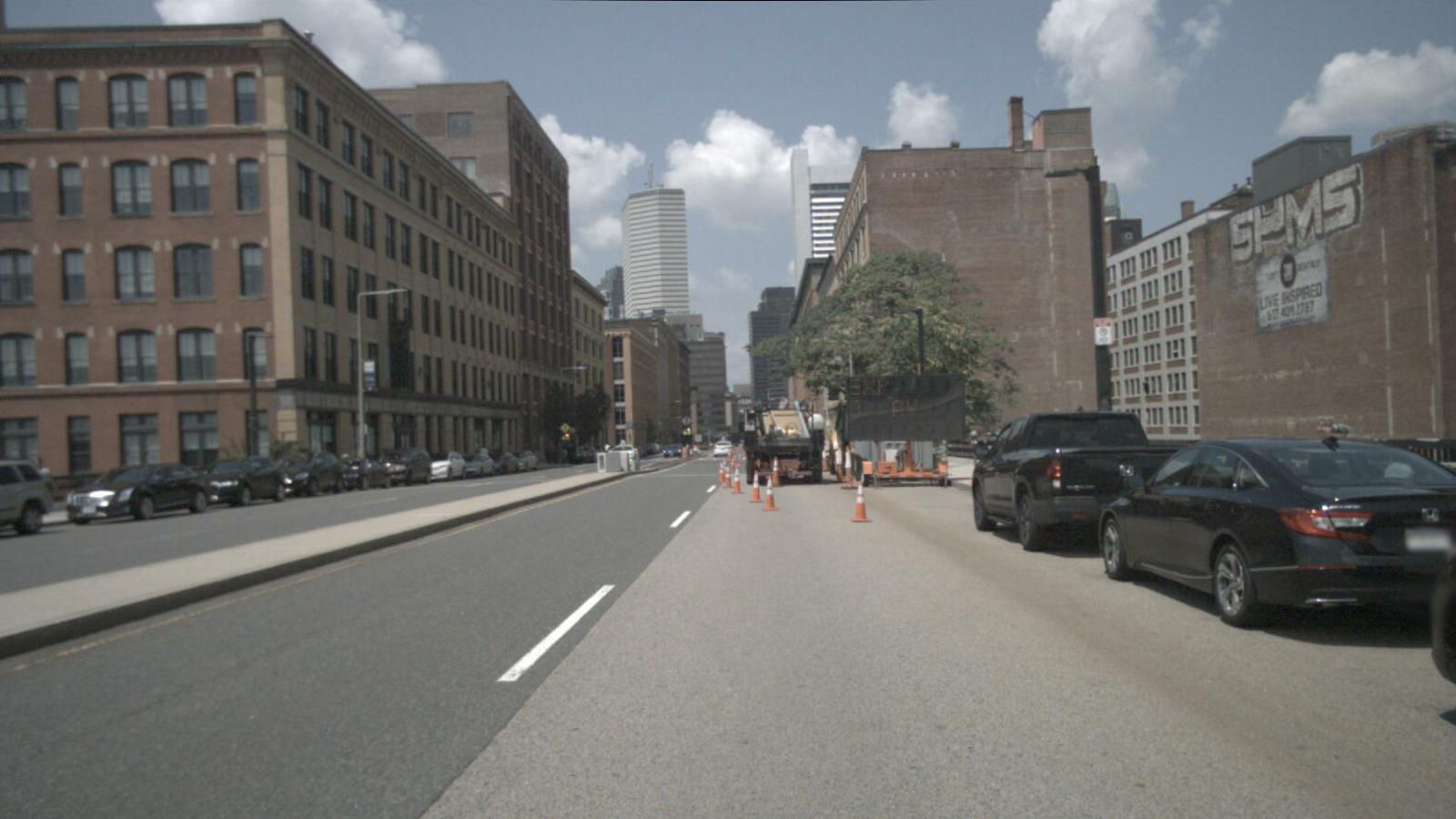}
\includegraphics[width=0.48\linewidth]{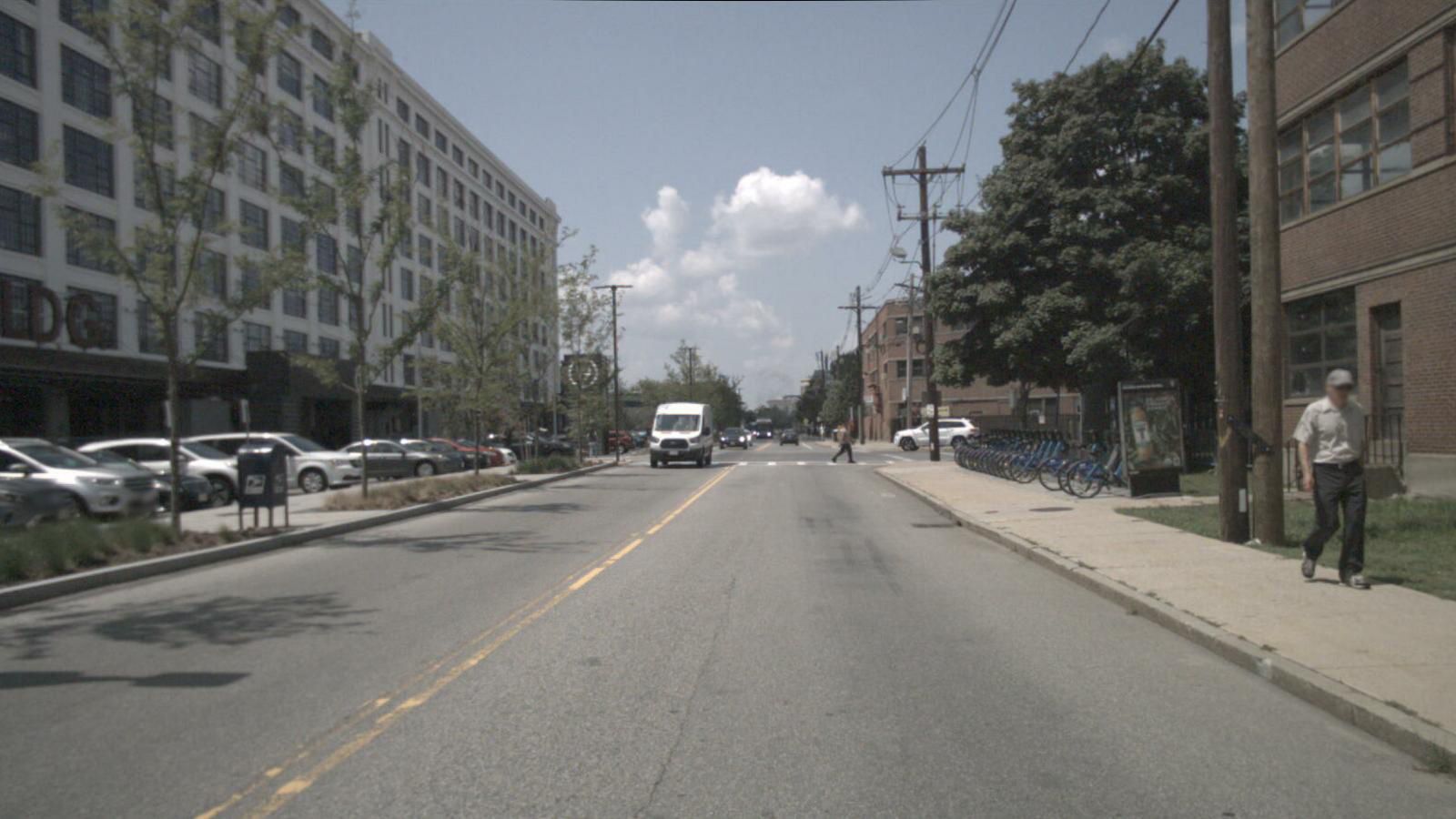}
\includegraphics[width=0.48\linewidth]{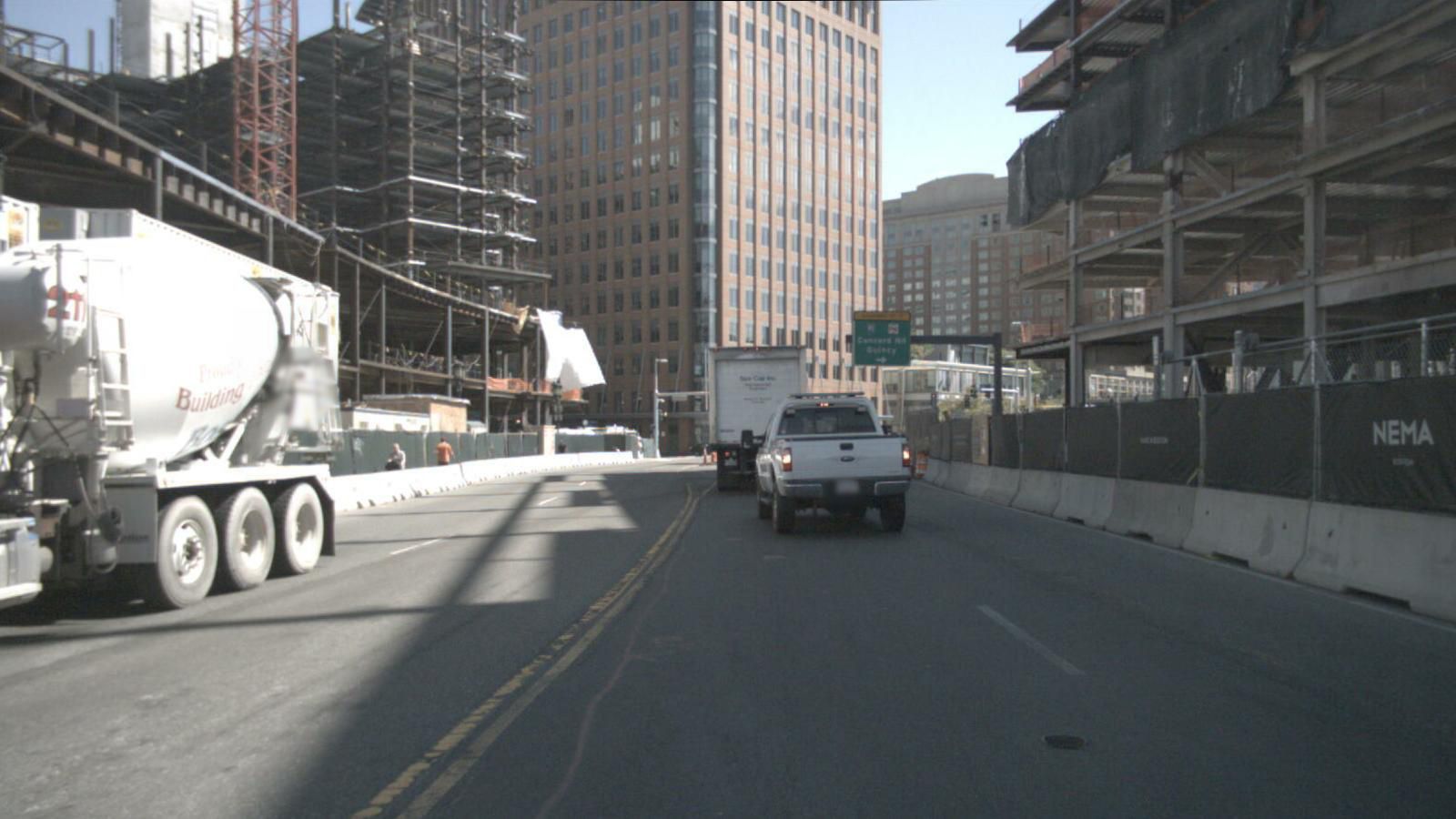}
    \includegraphics[width=0.48\linewidth]{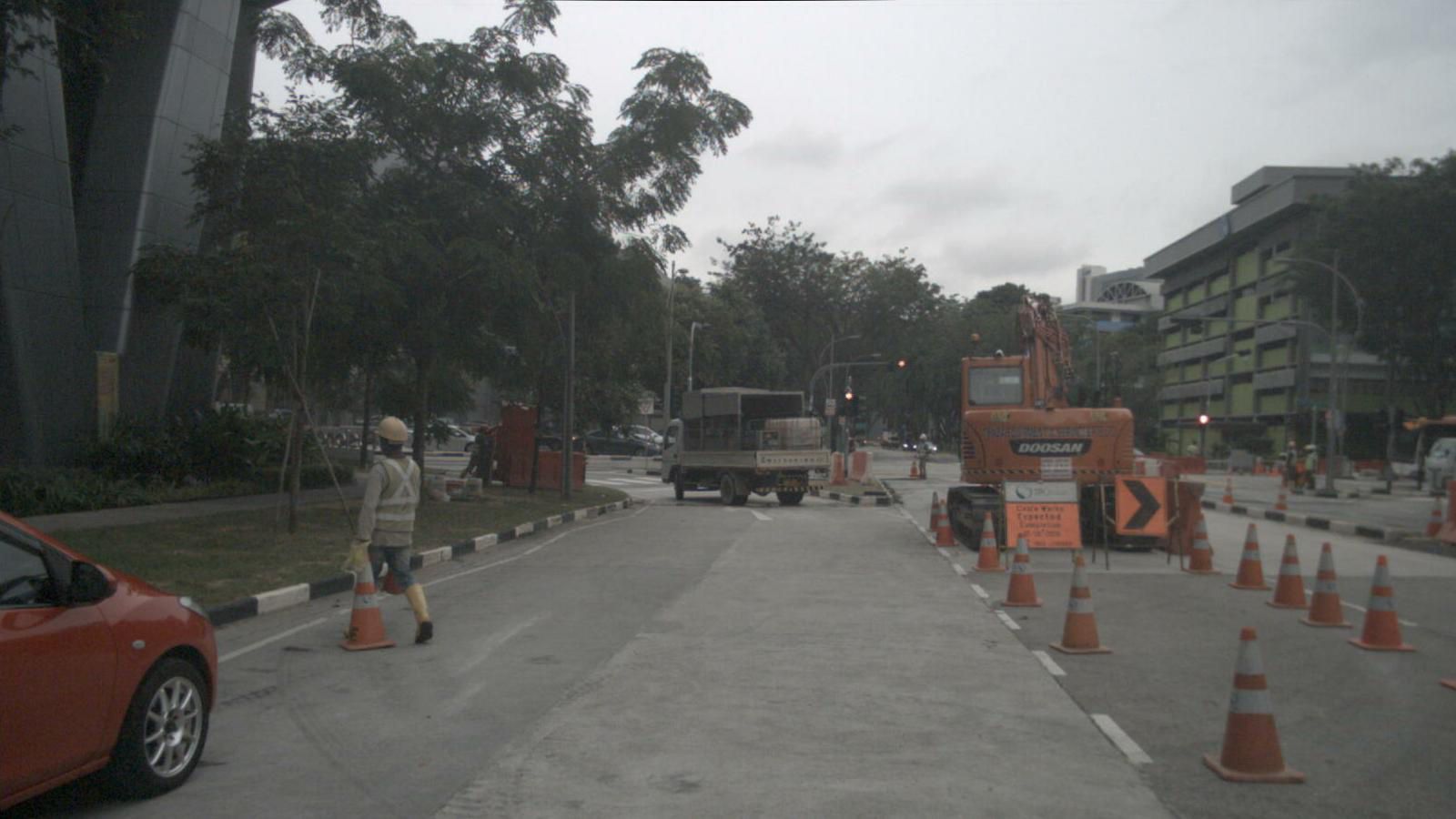}
\caption{\small Targeted or detected rare training data examples: Scenes containing rare objects such as motorcycles, construction vehicles and bicycles identified by the concept-based detection system. Targeting these examples significantly improved the detection performance for safety-critical categories. See Appendix Figure~\ref{fig:training_set_examples} for more images.}
\label{fig:additional_examples}
\end{figure}

Figure \ref{fig:additional_examples} presents several examples of targeted or detected rare training data scenes identified by our concept-based detection system.
These images showcase scenes that contain safety-critical rare objects such as motorcycles and bicycles, which were successfully identified and included in the training dataset. 
The ability to accurately extract these critical scenes is a significant advantage of our approach. When we target these examples, detection performance for safety-critical categories significantly improves, as demonstrated in our quantitative results in Table \ref{tab:results}. 
The successful extraction of rare and challenging scenes validates the practical value of our concept-based data-mining framework for real-world autonomous driving applications.

\section{Conclusion}
We presented a concept-based explainable data mining framework that leverages Vision-Language Models to improve 3D object detection performance. Our approach combines object detection, feature extraction, concept analysis, and outlier detection to identify valuable rare objects for inclusion in the training dataset.

This work makes several key contributions: \begin{itemize}
    \item \textbf{Explainable Data Mining Framework} that provides transparent, concept-based explanations for data selection decisions, which enhances trust and interpretability.
    \item \textbf{Cross-Modal Integration} that bridges 2D vision-language understanding with 3D detection through foundation models.
    \item \textbf{Efficient Performance Improvement} with significant gains for rare object categories while using only 20\% of the training data.
    \item \textbf{Practical Safety Contributions} through targeted improvements for safety-critical objects.
    \item \textbf{Annotation Cost Reduction} by identifying scenes with rare concepts such as work zones and emergency vehicles.
\end{itemize}

Explainability is a central strength of the framework, as demonstrated through multiple complementary approaches. VLM-generated natural language captions provide semantic understanding, while concept similarity visualizations (Figures \ref{fig:construction}, \ref{fig:motorcycle}, \ref{fig:bicycle}) and outlier detection visualization in t-SNE space (Figure \ref{fig:tsne}) offer interpretable insights. This explainability enables users to understand why specific objects are flagged as rare or important. These features create a comprehensive system that transparently improves the detection performance, which is essential for developing reliable perception systems for critical driving scenarios.

The results confirm that intelligent data selection through concept-based analysis can be more effective than simply increasing data quantity. This has important implications for dataset curation in autonomous driving, where obtaining diverse and representative datasets is often challenging and expensive. By enabling the identification of the most informative examples for model training, our approach addresses the need to handle rare but critical scenarios that may have safety-related implications in the future.
The cross-modal nature of our framework leverages 2D vision-language understanding to enhance 3D detection, bridging these historically separate domains. As VLMs continue to advance, our framework can benefit from these improvements without requiring architectural changes, making it adaptable to future developments.

Despite its strengths, the current study does not systematically analyze the sensitivity of the outlier detectors and their hyperparameters. In addition, it does not provide direct comparisons with alternative data selection strategies such as active learning or hard example mining. We will address these issues in our future work.

Future work could also explore the integration of this approach with active learning pipelines, extend it to video sequences for temporal consistency, and apply it to other perception tasks such as semantic segmentation and tracking. Additionally, applying this framework to real-world datasets beyond nuScenes would enable the targeted mining of rare objects in diverse environments, with the potential to reduce the annotation effort by selectively labeling the most informative instances.

\section*{Acknowledgements}
I am deeply grateful to Tier IV Inc. for its generous support throughout this study. I would like to express my sincere appreciation to all the current and former members of the company for providing essential computational resources and insightful guidance that greatly contributed to this study. Their technical expertise, openness to discussion, and continued encouragement have been invaluable to the progress of this research.
This paper is based on the work that I conducted during my internship at Tier IV Inc. from October 2024 to April 2025.
\bibliographystyle{plain} 
\bibliography{bmvc_style_references}

@inproceedings{lang2019pointpillars,
  title={PointPillars: Fast Encoders for Object Detection from Point Clouds},
  author={Lang, Alex H. and Vora, Sourabh and Caesar, Holger and Zhou, Lubing and Yang, Jiong and Beijbom, Oscar},
  booktitle={Proceedings of the IEEE/CVF Conference on Computer Vision and Pattern Recognition (CVPR)},
  pages={12697--12705},
  year={2019}
}

@InProceedings{yin2021center,
    author    = {Yin, Tianwei and Zhou, Xingyi and Krahenbuhl, Philipp},
    title     = {Center-Based 3D Object Detection and Tracking},
    booktitle = {Proceedings of the IEEE/CVF Conference on Computer Vision and Pattern Recognition (CVPR)},
    month     = {June},
    year      = {2021},
    pages     = {11784-11793}
}

@inproceedings{bai2022transfusion,
  title={TransFusion: Robust LiDAR-Camera Fusion for 3D Object Detection with Transformers},
  author={Bai, Xuyang and Hu, Zeyu and Zhu, Xinge and Huang, Qingqiu and Chen, Yilun and Fu, Hongbo and Tai, Chiew-Lan},
  booktitle={Proceedings of the IEEE/CVF Conference on Computer Vision and Pattern Recognition (CVPR)},
  pages={1090--1099},
  year={2022}
}

@inproceedings{liu2023bevfusion,
  title={BEVFusion: Multi-Task Multi-Sensor Fusion with Unified Bird's-Eye View Representation},
  author={Liu, Zhijian and Tang, Haotian and Amini, Alexander and Yang, Xinyu and Mao, Huizi and Rus, Daniela L and Han, Song},
  booktitle={2023 IEEE International Conference on Robotics and Automation (ICRA)},
  pages={2774--2781},
  year={2023},
  organization={IEEE}
}

@inproceedings{radford2021learning,
  title={Learning transferable visual models from natural language supervision},
  author={Radford, Alec and Kim, Jong Wook and Hallacy, Chris and Ramesh, Aditya and Goh, Gabriel and Agarwal, Sandhini and Sastry, Girish and Askell, Amanda and Mishkin, Pamela and Clark, Jack and others},
  booktitle={International conference on machine learning},
  pages={8748--8763},
  year={2021},
  organization={PMLR}
}

@inproceedings{li2022blip,
  title={Blip: Bootstrapping language-image pre-training for unified vision-language understanding and generation},
  author={Li, Junnan and Li, Dongxu and Xiong, Caiming and Hoi, Steven},
  booktitle={International conference on machine learning},
  pages={12888--12900},
  year={2022},
  organization={PMLR}
}

@inproceedings{liu2024grounding,
  title={Grounding dino: Marrying dino with grounded pre-training for open-set object detection},
  author={Liu, Shilong and Zeng, Zhaoyang and Ren, Tianhe and Li, Feng and Zhang, Hao and Yang, Jie and Jiang, Qing and Li, Chunyuan and Yang, Jianwei and Su, Hang and others},
  booktitle={European Conference on Computer Vision},
  pages={38--55},
  year={2024}
}

@article{liu2023visual,
  title={Visual instruction tuning},
  author={Liu, Haotian and Li, Chunyuan and Wu, Qingyang and Lee, Yong Jae},
  journal={Advances in neural information processing systems},
  volume={36},
  pages={34892--34916},
  year={2023}
}

@inproceedings{jiang2022improving,
  title={Improving the Intra-class Long-tail in 3D Detection via Rare Example Mining},
  author={Jiang, Chiyu Max and Najibi, Mahyar and Qi, Charles R. and Zhou, Yin and Anguelov, Dragomir},
  booktitle={Proceedings of the European Conference on Computer Vision (ECCV)},
  pages={158--175},
  year={2022}
}

@inproceedings{peri2023towards,
  title={Towards Long-Tailed 3D Detection},
  author={Peri, Neehar and Dave, Achal and Ramanan, Deva and Kong, Shu},
  booktitle={Conference on Robot Learning},
  pages={1904--1915},
  year={2023},
  organization={PMLR}
}

@inproceedings{lu2023open,
  title={Open-vocabulary point-cloud object detection without 3D annotation},
  author={Lu, Yuheng and Xu, Chenfeng and Wei, Xiaobao and Xie, Xiaodong and Tomizuka, Masayoshi and Keutzer, Kurt and Zhang, Shanghang},
  booktitle={Proceedings of the IEEE/CVF conference on computer vision and pattern recognition},
  pages={1190--1199},
  year={2023}
}

@inproceedings{ye2025vlmine,
  title={Vlmine: Long-tail data mining with vision language models},
  author={Ye, Mao and Meyer, Greg P and Zhang, Zaiwei and Park, Dennis and Mustikovela, Siva Karthik and Chai, Yuning and Wolff, Eric},
  booktitle={Proceedings of the Winter Conference on Applications of Computer Vision},
  pages={1072--1082},
  year={2025}
}

@inproceedings{caesar2020nuscenes,
  title={nuScenes: A Multimodal Dataset for Autonomous Driving},
  author={Caesar, Holger and Bankiti, Varun and Lang, Alex H. and Vora, Sourabh and Liong, Venice Erin and Xu, Qiang and Krishnan, Anush and Pan, Yu and Baldan, Giancarlo and Beijbom, Oscar},
  booktitle={Proceedings of the IEEE/CVF Conference on Computer Vision and Pattern Recognition (CVPR)},
  pages={11621--11631},
  year={2020}
}

@article{oikarinen2023label,
  title={Label-free concept bottleneck models},
  author={Oikarinen, Tuomas and Das, Subhro and Nguyen, Lam M and Weng, Tsui-Wei},
  journal={arXiv preprint arXiv:2304.06129},
  year={2023}
}

@article{cui2023ceir,
  title={CEIR: Concept-based Explainable Image Representation Learning},
  author={Cui, Yan and Liu, Shuhong and Li, Liuzhuozheng and Yuan, Zhiyuan},
  journal={arXiv preprint arXiv:2312.10747},
  year={2023}
}

@inproceedings{bengio2009curriculum,
  title={Curriculum learning},
  author={Bengio, Yoshua and Louradour, J{\'e}r{\^o}me and Collobert, Ronan and Weston, Jason},
  booktitle={Proceedings of the 26th annual international conference on machine learning},
  pages={41--48},
  year={2009}
}

@inproceedings{shrivastava2016training,
  title={Training Region-Based Object Detectors with Online Hard Example Mining},
  author={Shrivastava, Abhinav and Gupta, Abhinav and Girshick, Ross},
  booktitle={Proceedings of the IEEE Conference on Computer Vision and Pattern Recognition (CVPR)},
  pages={761--769},
  year={2016}
}

@article{maaten2008visualizing,
  title={Visualizing data using t-SNE},
  author={Maaten, Laurens van der and Hinton, Geoffrey},
  journal={Journal of machine learning research},
  volume={9},
  number={Nov},
  pages={2579--2605},
  year={2008}
}

@inproceedings{liu2008isolation,
  title={Isolation Forest},
  author={Liu, Fei Tony and Ting, Kai Ming and Zhou, Zhi-Hua},
  booktitle={2008 Eighth IEEE International Conference on Data Mining},
  pages={413--422},
  year={2008}
}

@article{ge2021yolox,
  title={YOLOX: Exceeding YOLO Series in 2021},
  author={Ge, Zheng and Liu, Songtao and Wang, Feng and Li, Zeming and Sun, Jian},
  journal={arXiv preprint arXiv:2107.08430},
  year={2021}
}

@article{reis2023realtime,
  title={Real-Time Flying Object Detection with YOLOv8},
  author={Reis, Dillon and Kupec, Jordan and Hong, Jacqueline and Daoudi, Ahmad},
  journal={arXiv preprint arXiv:2305.09972},
  year={2023}
}

@article{wang2024qwen2,
  title={Qwen2-VL: Enhancing vision-language model's perception of the world at any resolution},
  author={Wang, Peng and Bai, Shuai and Tan, Sinan and Wang, Shijie and Fan, Zhihao and Bai, Jinze and Chen, Keqin and Liu, Xuejing and Wang, Jialin and Ge, Wenbin and others},
  journal={arXiv preprint arXiv:2409.12191},
  year={2024}
}

@article{mao20233d,
  title={3D object detection for autonomous driving: A comprehensive survey},
  author={Mao, Jiageng and Shi, Shaoshuai and Wang, Xiaogang and Li, Hongsheng},
  journal={International Journal of Computer Vision},
  volume={131},
  number={8},
  pages={1909--1963},
  year={2023},
  publisher={Springer}
}

@article{guo2020deep,
  title={Deep Learning for 3D Point Clouds: A Survey},
  author={Guo, Yulan and Wang, Hanyun and Hu, Qingyong and Liu, Hao and Liu, Li and Bennamoun, Mohammed},
  journal={IEEE Transactions on Pattern Analysis and Machine Intelligence},
  volume={43},
  number={12},
  pages={4338--4364},
  year={2020},
  publisher={IEEE}
}

@article{QIAN2022108524,
  title={BADet: Boundary-Aware 3D Object Detection from Point Clouds},
  journal={Pattern Recognition},
  volume={125},
  pages={108524},
  year={2022},
  issn={0031-3203},
  doi={https://doi.org/10.1016/j.patcog.2022.108524},
  author={Qian, Rui and Lai, Xin and Li, Xirong}
}

@article{bommasani2021opportunities,
  title={On the opportunities and risks of foundation models},
  author={Bommasani, Rishi and Hudson, Drew A and Adeli, Ehsan and Altman, Russ and Arora, Simran and von Arx, Sydney and Bernstein, Michael S and Bohg, Jeannette and Bosselut, Antoine and Brunskill, Emma and others},
  journal={arXiv preprint arXiv:2108.07258},
  year={2021}
}

@inproceedings{kojima2022large,
  title={Large Language Models are Zero-Shot Reasoners},
  author={Kojima, Takeshi and Gu, Shixiang Shane and Reid, Machel and Matsuo, Yutaka and Iwasawa, Yusuke},
  booktitle={Advances in Neural Information Processing Systems (NeurIPS)},
  year={2022}
}

@inproceedings{lin2018focal,
  title={Focal Loss for Dense Object Detection},
  author={Lin, Tsung-Yi and Goyal, Priya and Girshick, Ross and He, Kaiming and Doll{\'a}r, Piotr},
  booktitle={Proceedings of the IEEE International Conference on Computer Vision (ICCV)},
  pages={2980--2988},
  year={2017}
}

@inproceedings{koh2020concept,
  title={Concept bottleneck models},
  author={Koh, Pang Wei and Nguyen, Thao and Tang, Yew Siang and Mussmann, Stephen and Pierson, Emma and Kim, Been and Liang, Percy},
  booktitle={International conference on machine learning},
  pages={5338--5348},
  year={2020},
  organization={PMLR}
}

@article{achiam2023gpt,
  title={GPT-4 technical report},
  author={Achiam, Josh and Adler, Steven and Agarwal, Sandhini and Ahmad, Lama and Akkaya, Ilge and Aleman, Florencia Leoni and Almeida, Diogo and Altenschmidt, Janko and Altman, Sam and Anadkat, Shyamal and others},
  journal={arXiv preprint arXiv:2303.08774},
  year={2023}
}

\appendix
\begin{figure}[t]
\centering

\includegraphics[width=0.9\linewidth]{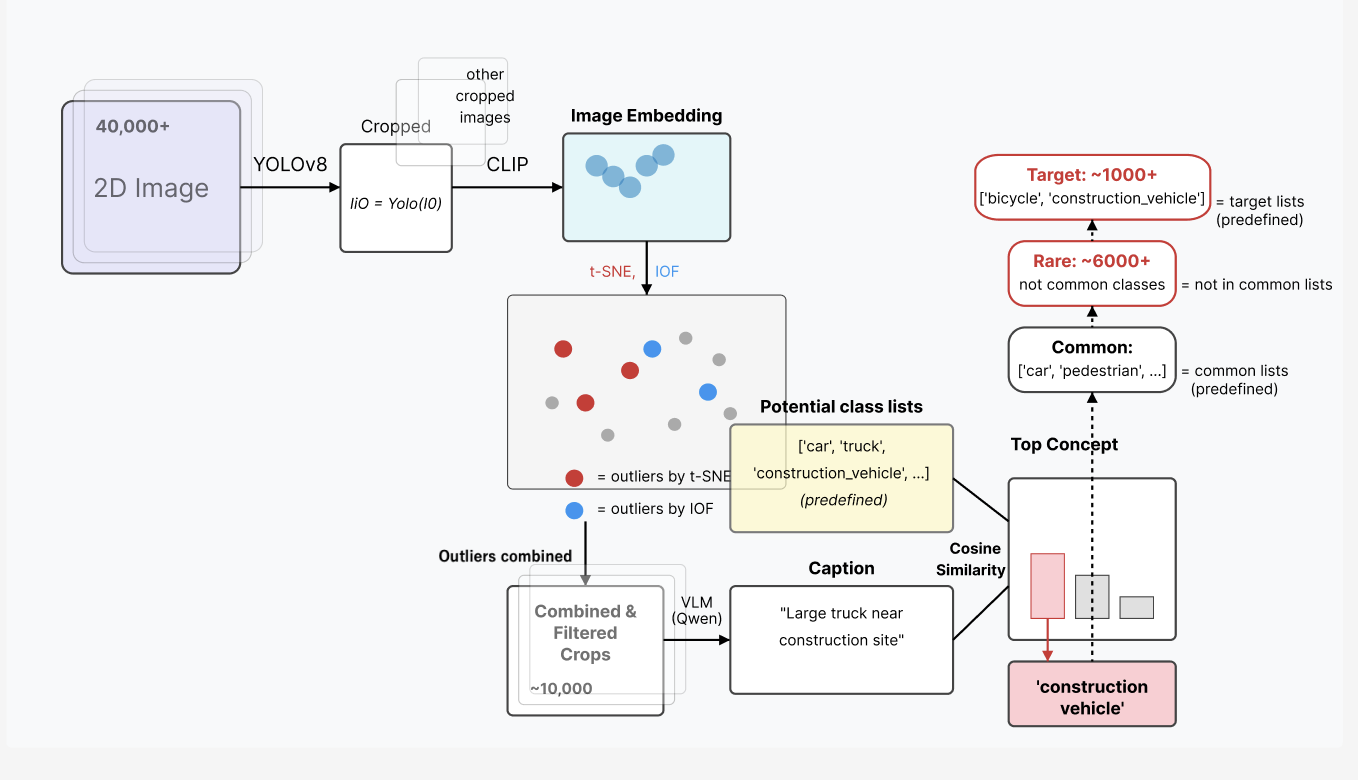}
\includegraphics[width=0.9\linewidth]{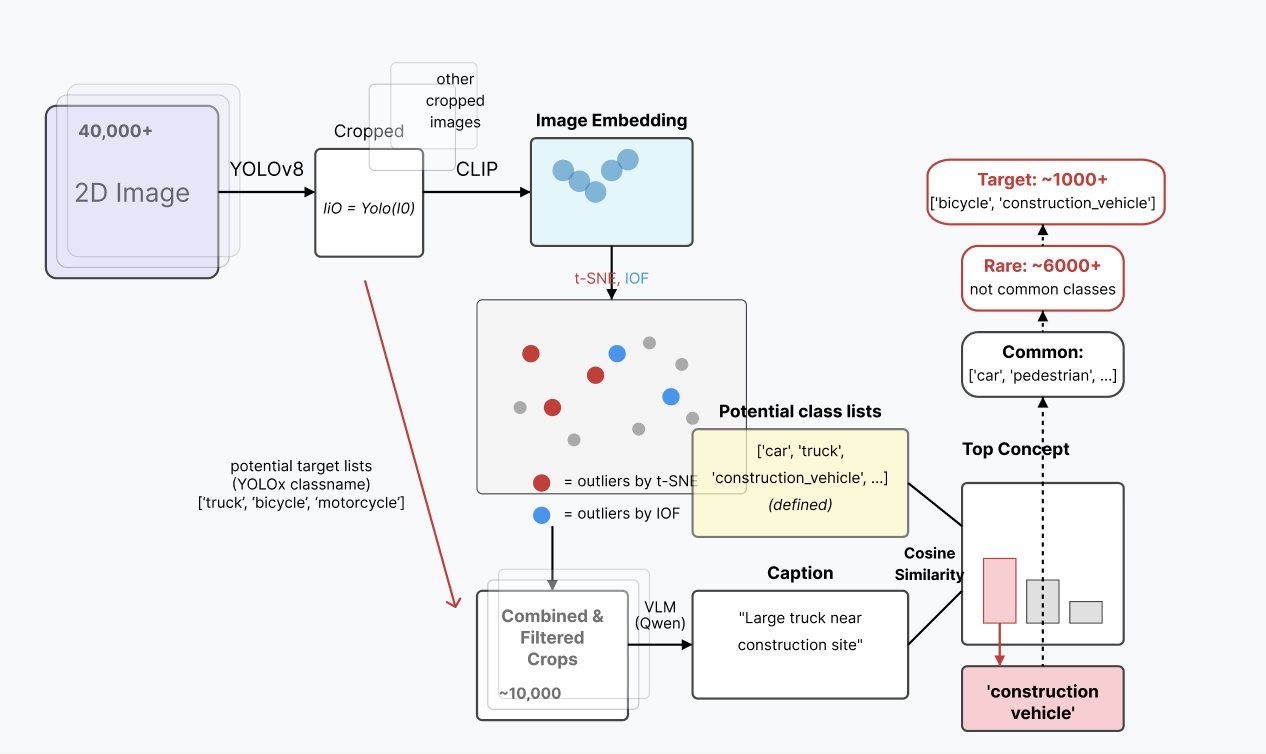}
\caption{\small 
Comparison of Random-Target (Top) and Random-Target+ (Bottom) approaches. Random-Target+ (Bottom) incorporates potential target objects into the caption generation pipeline regardless of their outlier status, indicated by the red arrow, to ensure comprehensive extraction of target objects such as trucks. Random-Target (Top) employs a more selective approach to target object inclusion.
} 
\label{fig:rare_target_flow_appendix}
\end{figure}
\section{Advanced Data Mining Approaches}
\label{sec:advanced_data_mining}

This section details the advanced data mining strategies we developed to improve detection of rare and safety-critical objects in autonomous driving scenarios.

\subsection{Random-Target and Random-Target+ Approaches}
\label{subsec:random_target_approaches}

We developed several specialized variants of our data mining framework to enhance coverage of rare and safety-critical objects:

\begin{itemize}
    \item \textbf{Random-Target (Bicycle, Motorcycle):} 
    This is the main method of this paper, and it focuses on two vulnerable road user categories that are critical for safety. By targeting both bicycles and motorcycles, it achieves enhanced performance in detecting these challenging categories while maintaining the same overall mAP improvement.

    \item \textbf{Random-Target (Bicycle, Construction Vehicle):} This setting targets bicycle and construction vehicle, which were the two categories with the lowest mAP in the Random 10\% experiment (Table~\ref{tab:extended_results}).
    \item \textbf{Random-Target+ (Bicycle, Construction Vehicle):} 
    This approach incorporates potential target objects into the caption generation pipeline regardless of their outlier status, indicated by the red arrow in Figure \ref{fig:rare_target_flow_appendix}. This ensures comprehensive extraction of near target objects such as construction vehicles and trucks. The approach is particularly effective for classes with high visual diversity, where standard outlier detection might miss important instances. The ‘truck’ class in YOLOX corresponds to all of ‘construction\_vehicle’, ‘truck’, and ‘trailer’ in nuScenes, and by including all of these near target objects in the caption generation process, this approach aims to capture most of the construction vehicles in nuscenes images.
\end{itemize}

\begin{table}[t]
\small 
\centering
\caption{\small 
Extended performance comparison on the nuScenes test set in terms of Average Precision (AP) for our advanced mining strategies. The Random-Target (B,CV) refers to bicycle and construction vehicle targeting, while Random-Target (B,M) refers to bicycle and motorcycle targeting. Asterisks (*) denotes cases where the accuracy improved by more than 0.5 AP compared to the Random 20\% setting.} 
\label{tab:extended_results}
\begin{adjustbox}{width=\textwidth}
\begin{tabular}{lcccccccccccc}
\toprule
\textbf{Data} & \textbf{mAP} & \textbf{Car} & \textbf{Truck} & \textbf{Bus} & \textbf{Tra} & \textbf{CV} & \textbf{Ped} & \textbf{Mot} & \textbf{Bic} & \textbf{Traf} & \textbf{Barrier} \\
\midrule
Random 20\% & 43.3 & 81.1 & 45.2 & 59.7 & 28.2 & 8.8 & 73.2 & 31.1 & 7.5 & 46.4 & 52.4 \\
Random-Rare & 44.0* & 81.4 & 46.1* & 59.9 & 30.5* & 9.4* & 73.0 & 31.2* & 7.8 & 48.5* & 52.4 \\
Random-Target (B,CV) & 44.7* & 81.3 & 45.4* & 58.1 & 32.3* & 8.5 & 73.1 & 32.7* & 11.9* & 49.4* & 54.2* \\
Random-Target (B,M) & 44.7* & 81.2 & 45.5* & 58.1 & 31.2* & 9.3* & 72.9 & 34.6* & 11.2* & 48.7* & 54.1* \\
Random-Target+ & 44.6* & 81.5 & 46.7* & 57.5 & 34.0* & 9.3* & 73.0 & 33.3* & 9.1* & 48.7* & 52.4 \\
\bottomrule
\end{tabular}
\end{adjustbox}
\end{table}

\subsection{Performance Analysis of Advanced Approaches}
\label{subsec:performance_analysis}

Performance results for our various Random-Target approaches are shown in Table \ref{tab:extended_results}. Notable observations include:

\begin{itemize}
    
    \item \textbf{Random-Target (Bicycle, Construction Vehicle):} Achieved exceptional performance in bicycle detection (11.9\%), demonstrating the benefit of selective filtering for bicycle objects. This approach shows strong trailer detection (32.3\%) but is less effective for construction vehicle detection compared to the Random-Rare approach.
    \item \textbf{Random-Target (Bicycle, Motorcycle):} By targeting two vulnerable road user categories, this approach achieved significant improvements in motorcycle detection (34.6\% vs 31.1\% for Random 20\%) while maintaining strong bicycle detection performance (11.2\% vs 7.5\% for Random 20\%). It also showed good performance in construction vehicle detection (9.3\%), suggesting that improved motorcycle detection indirectly benefits the detection of other rare categories. 
    \item \textbf{Random-Target+ (Bicycle, Construction Vehicle):} Achieved strong performance in truck detection (46.7\%) and trailer detection (34.0\%), outperforming the Random-Target approaches in these categories. This validates the effectiveness of including potential target objects regardless of outlier status for these visually diverse categories.

\end{itemize}

All variants of our approach demonstrate significant improvements over the Random 20\% baseline across most categories, with about the same 1.4 (or 1.3) mAP improvement (44.7\% vs 43.3\%). This confirms the effectiveness and flexibility of our concept-based data mining framework, which can be tailored to target different safety-critical object categories depending on specific application needs.

\subsubsection{Construction Vehicle Detection: Analysis and Challenges}
\label{subsubsec:construction_vehicle}

Construction vehicles present an interesting case study in our experiments. While the Random-Rare approach achieved slightly better performance (9.4\%) compared to the Random-Target (Bicycle, Construction Vehicle) approach (8.5\%), this can be explained by several factors:

\begin{itemize}
    \item \textbf{Inherent Rarity:} Construction vehicles are inherently rare in the nuScenes dataset, making them challenging targets for any data mining approach.
    
    \item \textbf{High Visual Diversity:} The construction vehicle category encompasses a wide variety of vehicle types (excavators, bulldozers, cement mixers, etc.) with significantly different visual appearances, making it difficult for VLMs to consistently identify all variants.
    
    \item \textbf{Visual Similarity to Trucks:} Many construction vehicles share visual characteristics with trucks, leading to potential confusion in VLM-based identification. Our analysis of concept similarity scores revealed frequent overlap between truck and construction vehicle categories.
\end{itemize}

Interestingly, the Random-Target (Bicycle, Motorcycle) approach achieved 9.3\% AP for construction vehicles despite not explicitly targeting this category. This suggests that, due to the high intra-class diversity of construction vehicles, a more generic rare-object targeting strategy can outperform explicitly targeting this category when the VLM’s recognition capability is not yet reliable. Furthermore, since the architecture allows for flexible replacement of the VLM component (validated with BLIP, Qwen-based models, and OpenAI-based VLMs), the AP for construction vehicles is expected to improve as VLM accuracy improves. This highlights the robustness and generalizability of our method.

\section{Implementation Details of the Concept-based Data Mining Framework}
\label{sec:implementation_details}

This section provides information about the implementation of our concept-based data mining framework, focusing on the algorithmic innovations that enable effective rare object detection.

\subsection{Multiple Outlier Detection Approaches}
\label{subsec:outlier_detection}

\begin{figure}[t]
    \centering
    \includegraphics[width=0.7\linewidth]{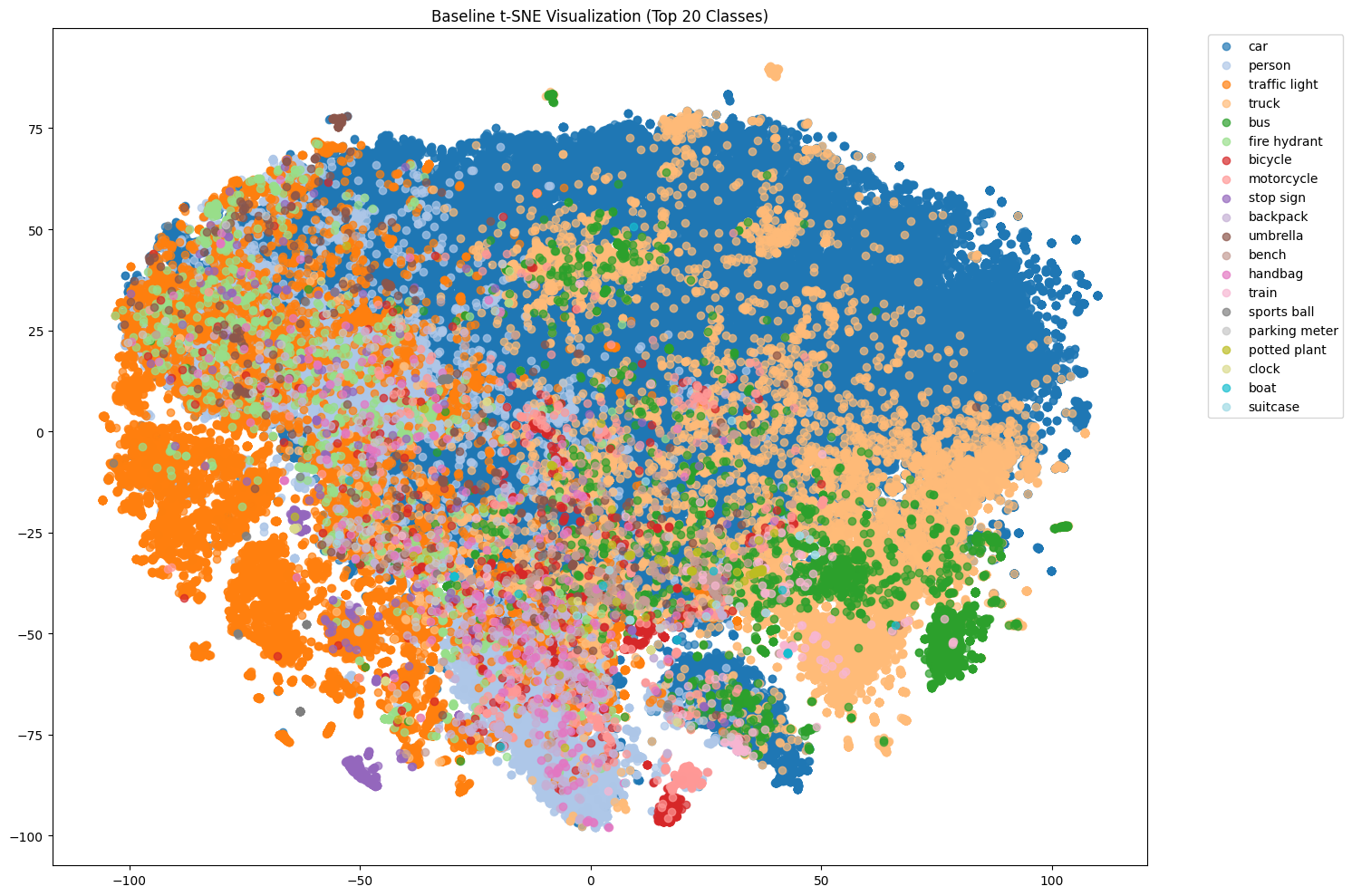}
    \includegraphics[width=0.7\linewidth]{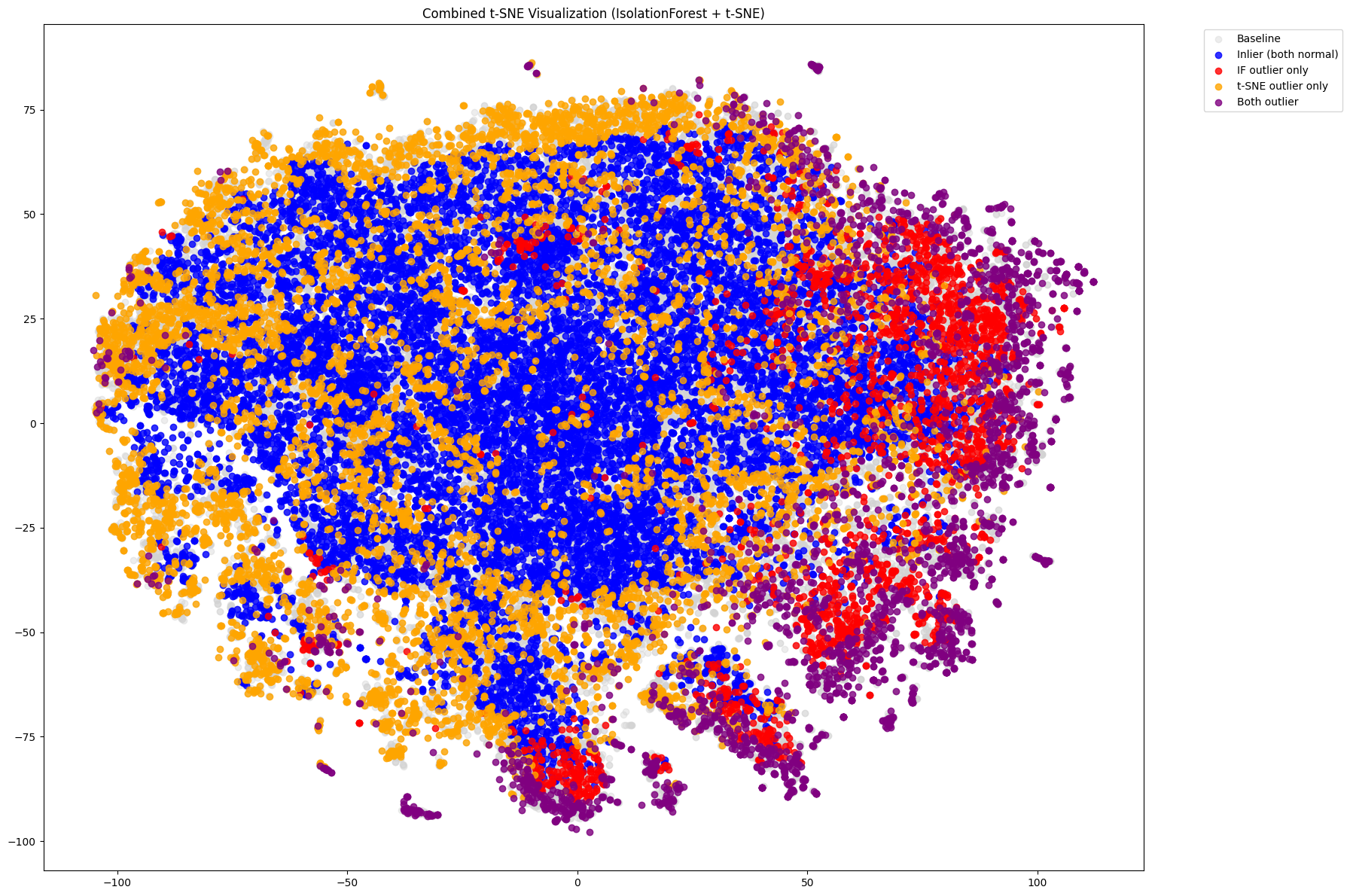}
    \caption{t-SNE visualization of object embeddings from our early experiment where we trained the Isolation Forest model on 10\% of the initial nuScenes dataset and attempted to detect outliers from the remaining 90\%. (\textit{Top}) Embeddings colored by object category. (\textit{Bottom}) Isolation Forest outlier detection overlaid on the t-SNE map: detected outliers are highlighted, while inliers retain their category-based colors. This approach helped establish the foundation for our current outlier detection methodology.}
    \label{fig:tsne_appendix}
\end{figure}

We tried a diverse set of outlier detection algorithms to enhance the robustness of rare object identification:

\begin{itemize}
    \item \textbf{Isolation Forest:} We utilize Isolation Forest with a contamination parameter of 0.20, which efficiently identifies non-linear anomalies by isolating observations through recursive partitioning. 
    \item \textbf{Train-test split for outlier detection} In our early experiments, we explored training the Isolation Forest and t-SNE model on 10\% of the initial nuScenes dataset and then using it to detect outliers from the remaining 90\%. This approach, visualized in Figure \ref{fig:tsne_appendix}, provided valuable insights into the distribution of rare objects and helped establish our current methodology.
    
    \item \textbf{Local Outlier Factor (LOF):} This density-based method identifies outliers by comparing the local density of a sample with the densities of its neighbors, effectively detecting outliers in regions of varying density.
    
    \item \textbf{t-SNE Distance-based Detection:} After dimensionality reduction using t-SNE, we implement a distance-based outlier detection approach that identifies points with anomalously large distances to their nearest neighbors.
    
    \item \textbf{Ensemble Combination:} combines the results from multiple algorithms using either union or intersection operations, significantly improving the precision of outlier detection. While we initially experimented with IOF+LOF combinations, our comparative analysis revealed that the t-SNE+IOF combination yielded the most effective results for our specific task, particularly in identifying visually distinct rare objects relevant to autonomous driving. This combination was therefore selected for our final implementation.
\end{itemize}

\subsection{Class-Aware Outlier Detection}
\label{subsec:class_aware_detection}

Class imbalance presents a significant challenge in autonomous driving datasets. To address this issue, the proposed method employs class-aware outlier detection, which processes each class independently. This targeted approach accounts for class-specific characteristics, proving particularly effective when dealing with highly skewed distributions typical in autonomous driving data. The contamination parameters for each class are fine-tuned according to their frequency and importance in the dataset, providing greater flexibility in controlling outlier detection sensitivity across different object categories.

\subsection{Multi-modal Feature Extraction}
\label{subsec:multimodal_extraction}

Central to the framework's capacity for identifying semantically meaningful rare objects is the integration of visual and linguistic features through several complementary techniques:

\begin{itemize}
    \item \textbf{CLIP Embeddings:} The system leverages CLIP (ViT-B/32) to extract image embeddings, yielding robust visual representations with rich semantic understanding derived from large-scale vision-language pretraining.
    
    \item \textbf{Caption Generation:} A flexible captioning mechanism alternates between Qwen2-VL and BLIP models depending on specific requirements and availability, maximizing the quality of generated descriptions.
    
    \item \textbf{Feature Integration:} By combining visual and textual information, the approach creates a comprehensive semantic representation of each object, capturing nuances that might be missed by single-modality systems.
    
    \item \textbf{Weighted Similarity Calculation:} An adjustable weighting parameter balances textual and visual similarity scores, allowing fine-tuning based on detection requirements for different object categories.
\end{itemize}

\section{Exploration and Development Process}
\label{sec:exploration}

This section details key insights from our exploration process and challenges we overcame in developing the concept-based data mining framework.

\subsection{Key Insights and Challenges}
\label{subsec:key_insights}

Several important insights emerged during our development process:

\noindent\textbf{Class Imbalance Insights:} We discovered that raw instance counts do not necessarily predict detection performance. Despite construction vehicles having more instances than motorcycles in the nuScenes dataset, their detection performance was significantly lower. This is due to the higher visual diversity within the construction vehicle category compared to motorcycles.


\noindent\textbf{Visual-Semantic Alignment:} Ensuring consistency between visual features and semantic concepts was challenging. We developed a specialized mapping function to bridge YOLO's detection categories with nuScenes classes to improve concept matching accuracy.

\subsection{Ablation Studies and Performance Analysis}
\label{subsec:ablation}

Table \ref{tab:ablation} presents a summary of our key experimental configurations and their corresponding performance metrics.

\begin{table}[h]
\centering
\caption{Ablation study of different data mining configurations. The training set for the final method (our proposed approach) was constructed by randomly selecting samples from the entire pool of discoveries made using IOF \& t-SNE combined detection unified with near target (possible target classes), along with all samples for which the top concept was either "construction vehicle" or "bicycle."}
\label{tab:ablation}
\begin{tabular}{lcc}
\hline
\textbf{Configuration} & \textbf{mAP}  \\
\hline
Random-only baseline (20\%) & 43.4  \\
IOF+LOF with random sampling & 43.7  \\
almost Final but use BLIP for captioning & 44.0  \\
Final method & 44.7 \\ 
\hline
\end{tabular}
\end{table}

\noindent\textbf{Best Performing Configuration:} Our most successful approach (achieving 44.7 mAP) first identified potential classes (e.g., truck, bicycle) using YOLOv8, then applied the t-SNE+IOF combination for outlier detection, which proved more effective than other combinations like IOF+LOF. Specifically, the Isolation Forest was applied to detect outliers within target categories, while t-SNE visualization enhanced the identification of visually distinct rare objects. Qwen2-VL was used to generate captions for these detected objects, and concept similarity matching identified the most relevant rare objects. This was complemented with construction vehicle neighborhood mining for additional focused improvement.

\section{Model and Implementation Details}
\label{sec:model_details}

The following table provides technical specifications of the models and key parameters used in our final implementation:

\begin{table}[h]
\centering
\caption{Model specifications and key parameters}
\label{tab:model_specs}
\begin{tabular}{ll}
\hline
\textbf{Component} & \textbf{Specification} \\
\hline
Object Detection & YOLOv8-l (80 classes, 640×640 input) \\
VLM for Captioning & Qwen2-VL-2B-Instruct (float16 precision) \\
Feature Extractor & CLIP-ViT-B/32 (LAION-2B trained) \\
Isolation Forest & contamination=0.20, n\_estimators=100 \\
LOF & n\_neighbors=20, contamination=0.20 \\
t-SNE & perplexity=30, n\_components=2 \\
Cosine Similarity & text weight=0.5, image weight=0.5 \\
Hardware & NVIDIA A100 40GB GPU \\
Processing Time & ~3.5 hours for full pipeline on nuScenes train split \\
\hline
\end{tabular}
\end{table}

\section{Training Curve Comparison}
\label{sec:training_curves}

\begin{figure}[h]
    \centering
    \includegraphics[width=0.95\linewidth]{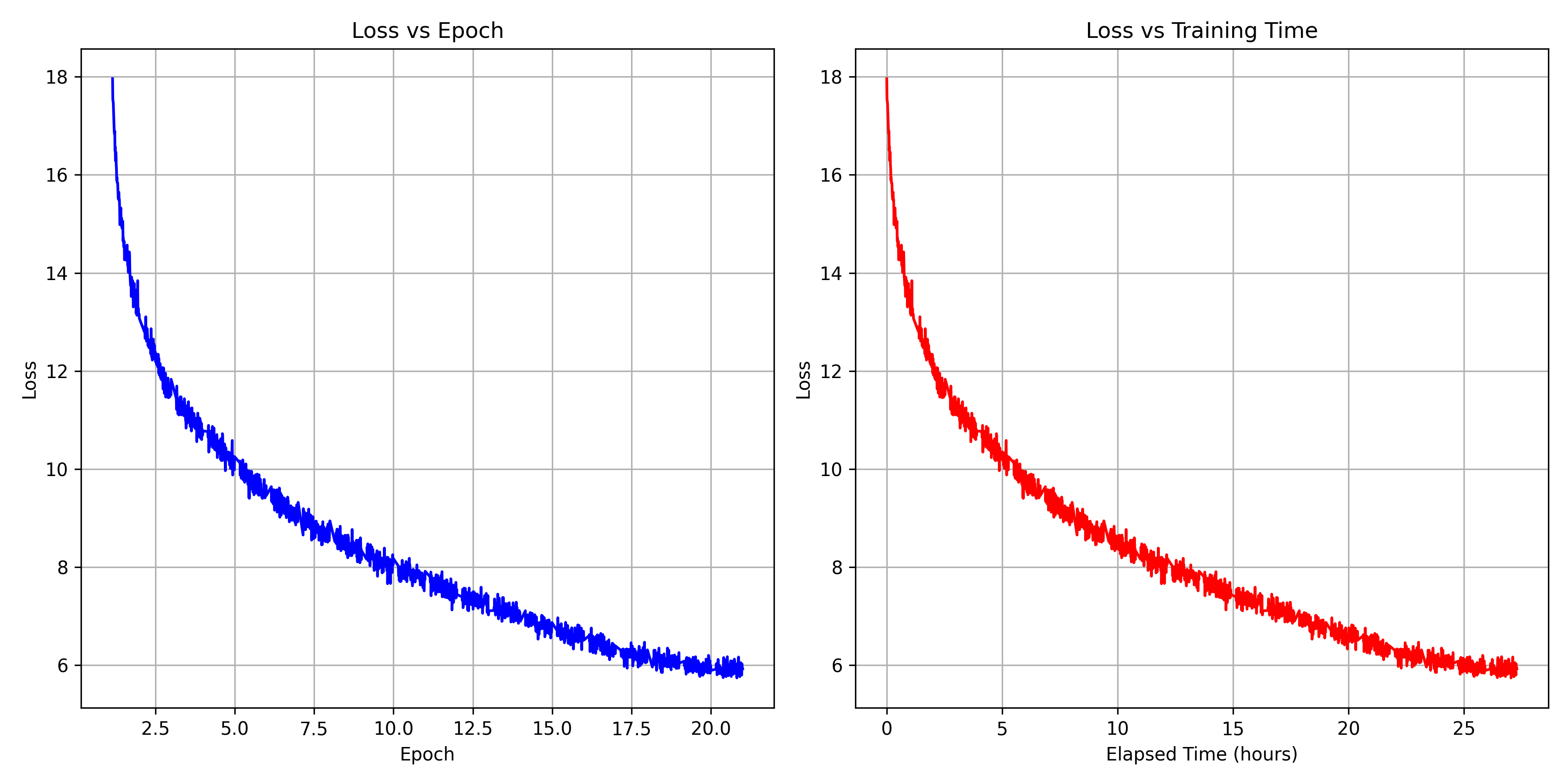}
    \includegraphics[width=0.9\linewidth]{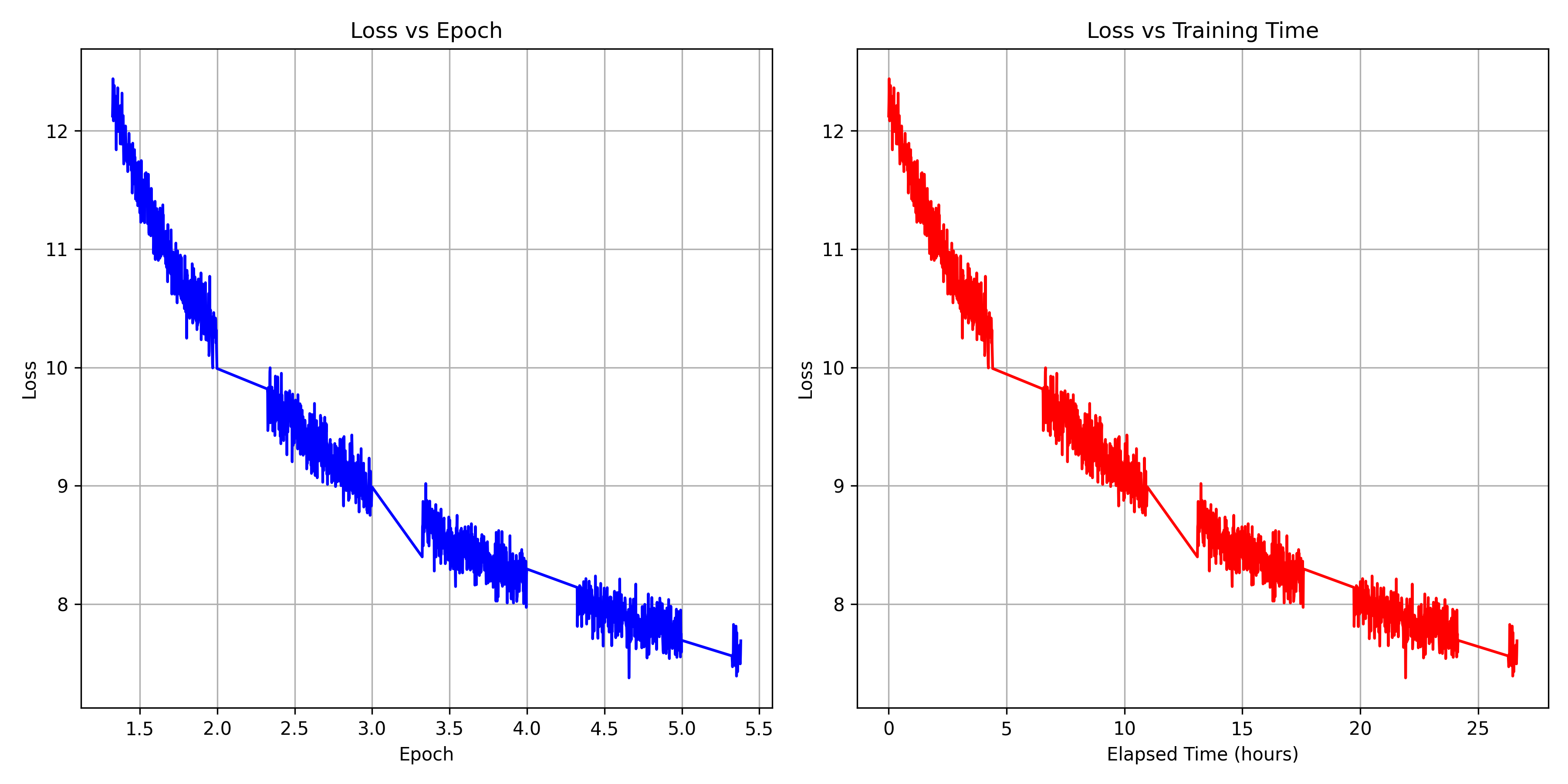}
    \caption{
        Training loss curves comparision for the proposed data mining strategy (Top) and baseline using whole training dataset (Bottom). The bottom plot is only about first 25 hours of training to compare the convergence speed.
        (\textbf{Left}) Loss vs Epoch. (\textbf{Right}) Loss vs Elapsed Training Time (hours).
        The plots show that not only does the final loss reach a lower value, but the convergence is also faster compared to the baseline.
         The bottom plot shows training results (for about first 25 hours) using the entire training set without data mining. When aligned by elapsed training time, the progress in terms of epochs is significantly slower, and the loss reduction is more gradual. This highlights the improved training efficiency enabled by our approach.
    }
    \label{fig:loss_curves}
\end{figure}

As shown in Figure~\ref{fig:loss_curves}, our data mining strategy enables faster convergence during training. The loss decreases more rapidly both in terms of epochs and actual elapsed training time, indicating improved sample efficiency and potential for reduced computational cost. This is an additional practical advantage, especially for large-scale or resource-constrained training scenarios.

\section{Training Set Examples}
\label{sec:training_examples}

\begin{figure}[h]
    \centering
    \includegraphics[width=0.48\linewidth]{Appendix/figs/n008-2018-07-27-12-07-38-0400__CAM_FRONT__1532708206512411.jpg}
    \includegraphics[width=0.48\linewidth]{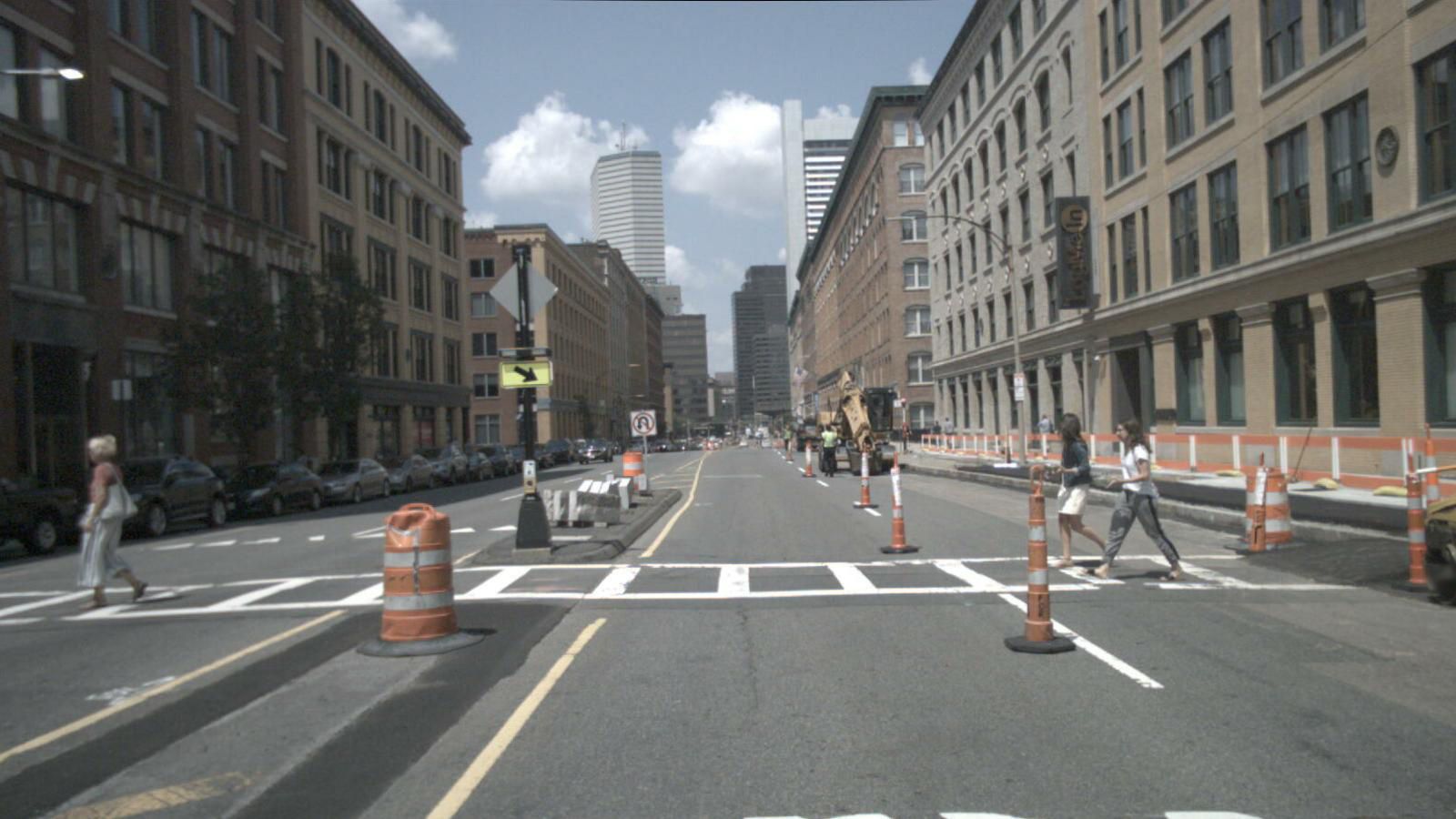}
    \includegraphics[width=0.48\linewidth]{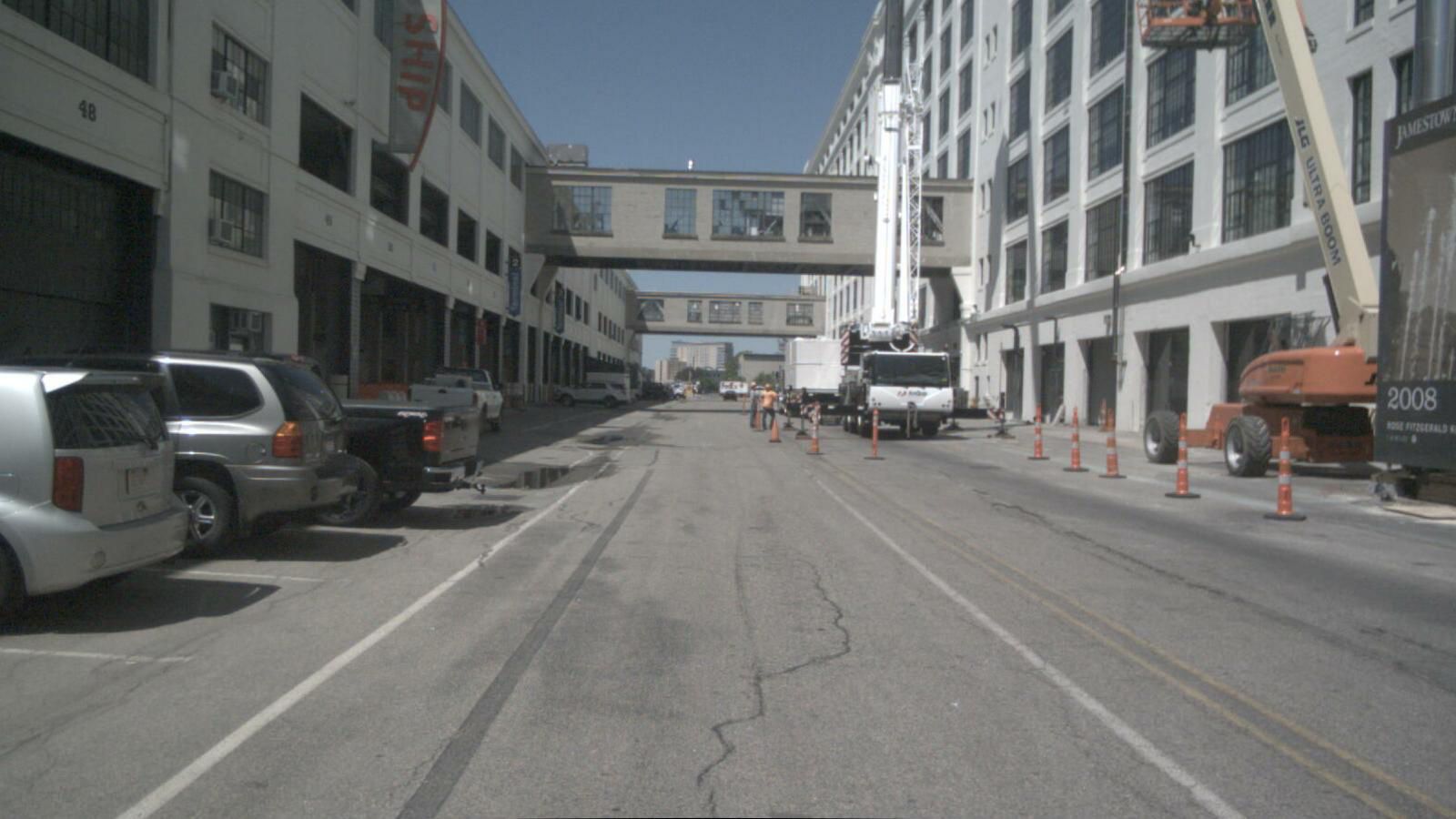}
    \includegraphics[width=0.48\linewidth]{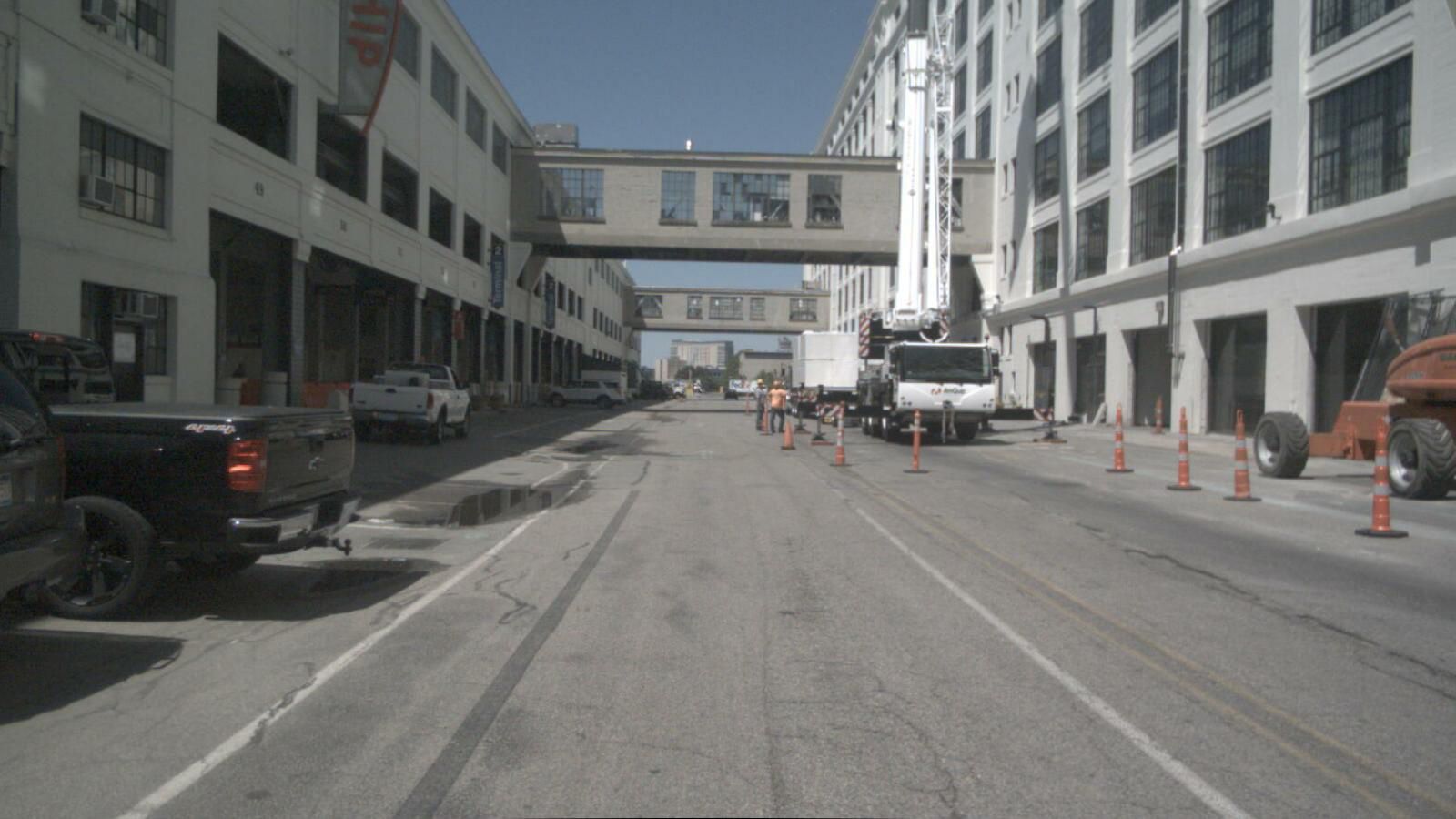}
    \includegraphics[width=0.48\linewidth]{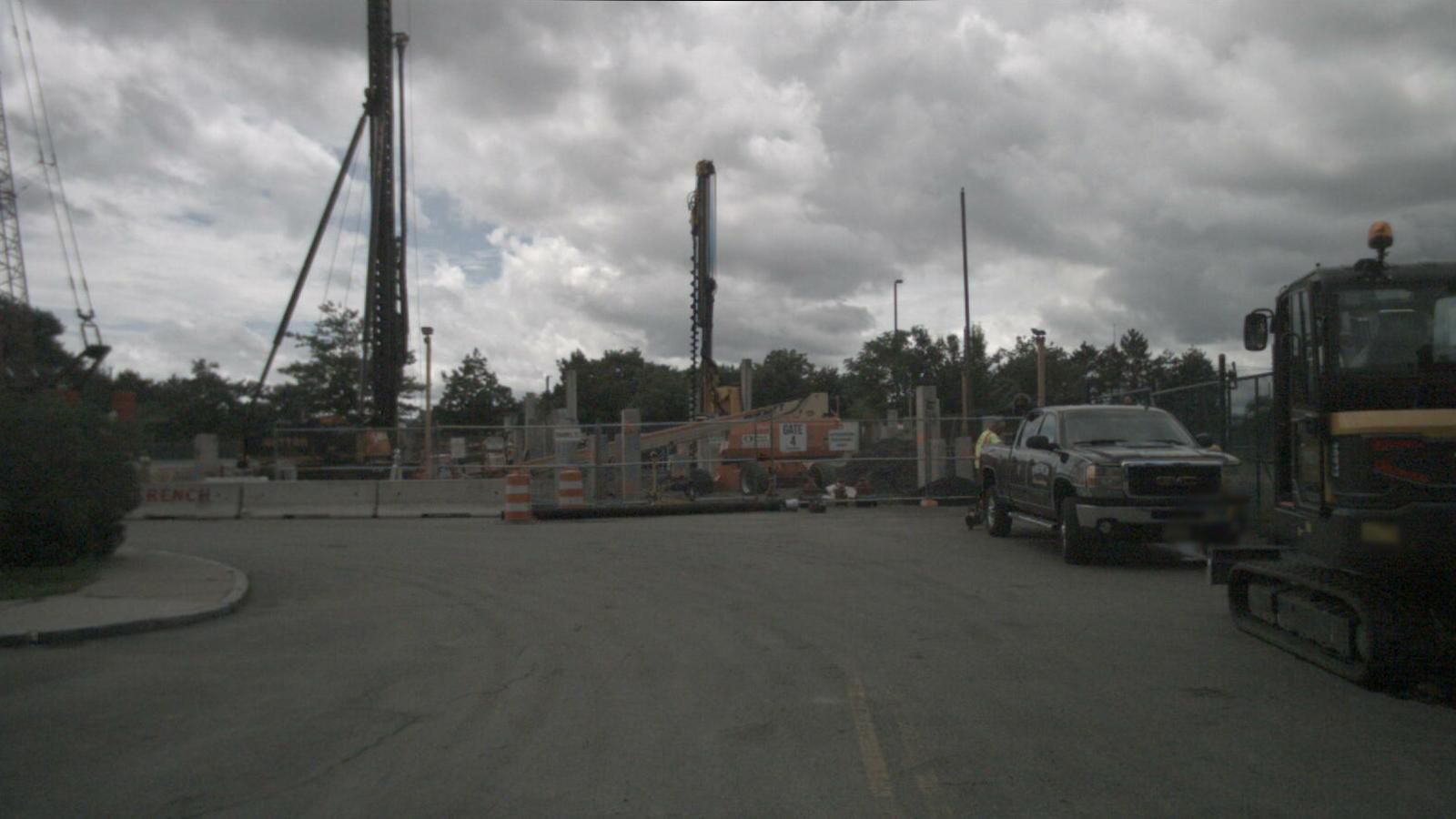}
    \includegraphics[width=0.48\linewidth]{Appendix/figs/n008-2018-07-27-12-07-38-0400__CAM_FRONT__1532707819862404.jpg}
    \includegraphics[width=0.48\linewidth]{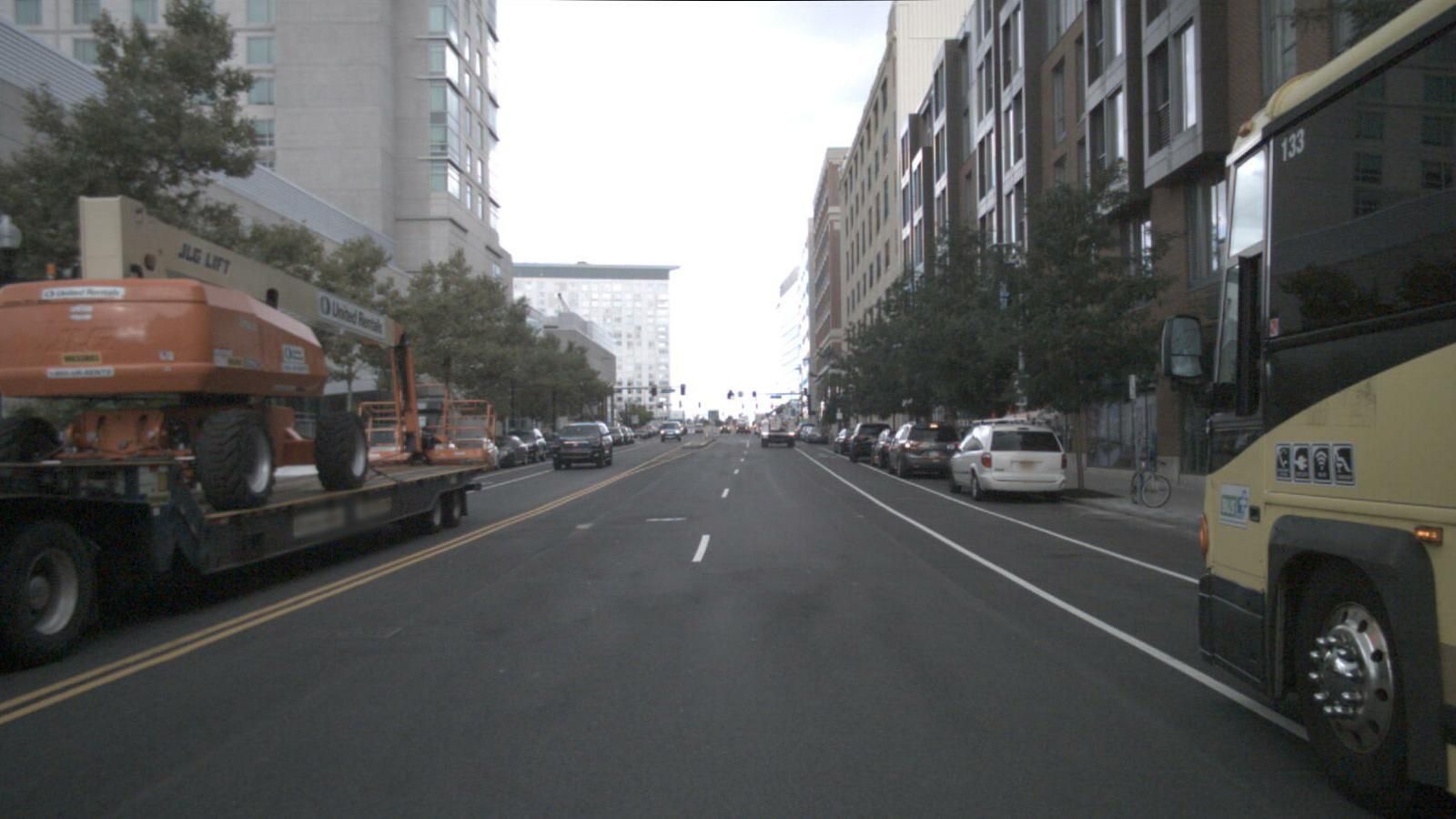}
    \includegraphics[width=0.48\linewidth]{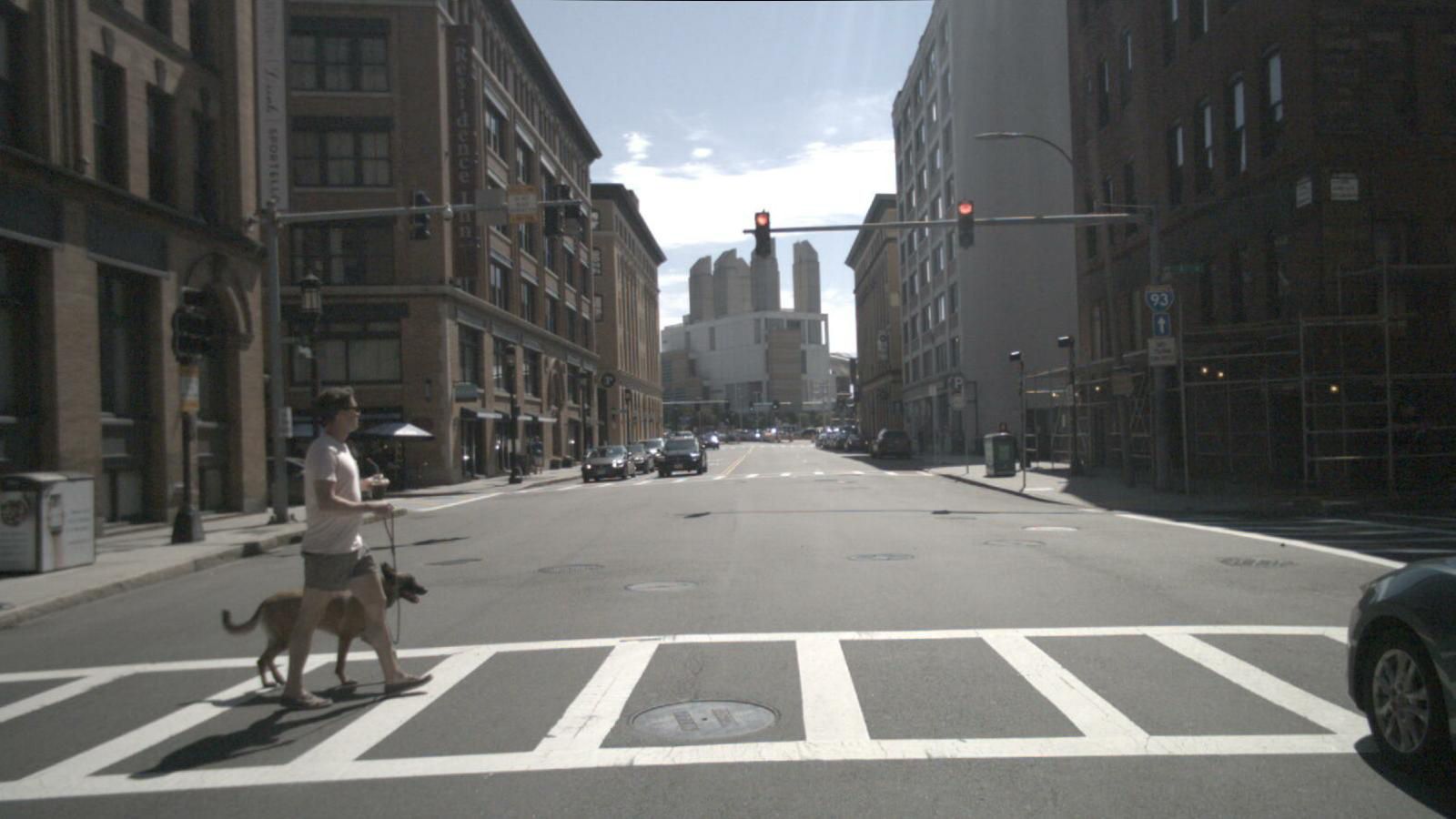}
    \caption{Training set examples including construction vehicles, bicycles, and other rare objects.}
    \label{fig:training_set_examples}
\end{figure}

This method constructs a training set that includes construction vehicles, bicycles, and other rare objects. A portion of the training set is shown in Figure~\ref{fig:training_set_examples}.

\end{document}